\documentclass[eng]{ajceam-class}

\usepackage{chemformula}
\usepackage{multirow}
\usepackage{algpseudocode}
\usepackage{float}
\usepackage{amsmath}
\usepackage{makecell}
\usepackage{listings}
\usepackage{stfloats}
\usepackage{hyperref}
\usepackage{afterpage}
\title{Structured information extraction from complex scientific text with fine-tuned large language models}

\author[1,2]{Alexander Dunn$^*$}
\author[1,2]{John Dagdelen$^*$}
\author[2]{Nicholas Walker}
\author[1,2]{Sanghoon Lee}
\author[2]{Andrew S. Rosen}
\author[1,3]{Gerbrand Ceder}
\author[1,2,4]{Kristin Persson}
\author[2]{Anubhav Jain}

\affil[1]{Materials Science and Engineering Department, University of California, Berkeley, California, USA}
\affil[2]{Energy Technologies Area, Lawrence Berkeley National Laboratory, California, USA}
\affil[3]{Materials Sciences Division, Lawrence Berkeley National Laboratory, California, USA}
\affil[4]{Molecular Foundry, Lawrence Berkeley National Laboratory, California, USA}

\firstauthor{SurnameA, SurnameB and SurnameC}

\contactauthor{Anubhav Jain}
\email{ajain@lbl.gov}

\thisvolume{XX}
\thisnumber{XX}
\thismonth{Month}
\thisyear{20XX}

\abstract{
Intelligently extracting and linking complex scientific information from unstructured text is a challenging endeavor particularly for those inexperienced with natural language processing. Here, we present a simple sequence-to-sequence approach to joint named entity recognition and relation extraction for complex hierarchical information in scientific text. The approach leverages a pre-trained large language model (LLM), GPT-3, that is fine-tuned on approximately 500 pairs of prompts (inputs) and completions (outputs). Information is extracted either from single sentences or across sentences in abstracts/passages, and the output can be returned as simple English sentences or a more structured format, such as a list of JSON objects. We demonstrate that LLMs trained in this way are capable of accurately extracting useful records of complex scientific knowledge for three representative tasks in materials chemistry: linking dopants with their host materials, cataloging metal-organic frameworks, and general chemistry/phase/morphology/application information extraction. This approach represents a simple, accessible, and highly-flexible route to obtaining large databases of structured knowledge extracted from unstructured text. An online demo is available at \url{http://www.matscholar.com/info-extraction}.}
 

\begin{document}

\maketitle
\printcontactdata

\section{Introduction} \label{sec:intro}

The majority of scientific knowledge about solid-state materials is scattered across the text, tables, and figures of millions of academic research papers. Thus, it is difficult for researchers to fully leverage existing knowledge when designing experiments or to even properly understand the full body of past work. While machine learning models for direct property prediction have been increasingly employed as screening steps for materials discovery and design workflows \cite{PropReviewSaal2020, Choudhary2022}, this approach is limited by the amount of training data available in tabulated databases. Such limitations are particularly apparent for experimental properties and parameters (in contrast to databases of materials property data derived from \textit{ab initio} simulations). Natural language processing (NLP) algorithms for materials, and named entity recognition (NER) models in particular, have made significant advances over the past half-decade towards structuring the existing body of textual materials science knowledge.\cite{Weston2019, Trewartha2022, Isazawa2022, Zhao2021} In materials NER models, entity labels such as "material" or "property" are applied to words in text and can be used, sometimes with additional post-processing, to construct auto-generated tabular databases of materials property data aggregated from text entries.\cite{Sierepeklis2022, Beard2022, Kumar2022, BatteryBERT, Dong2022}

\begin{figure*}[t]
    \centering
    \includegraphics[width=0.8\textwidth]{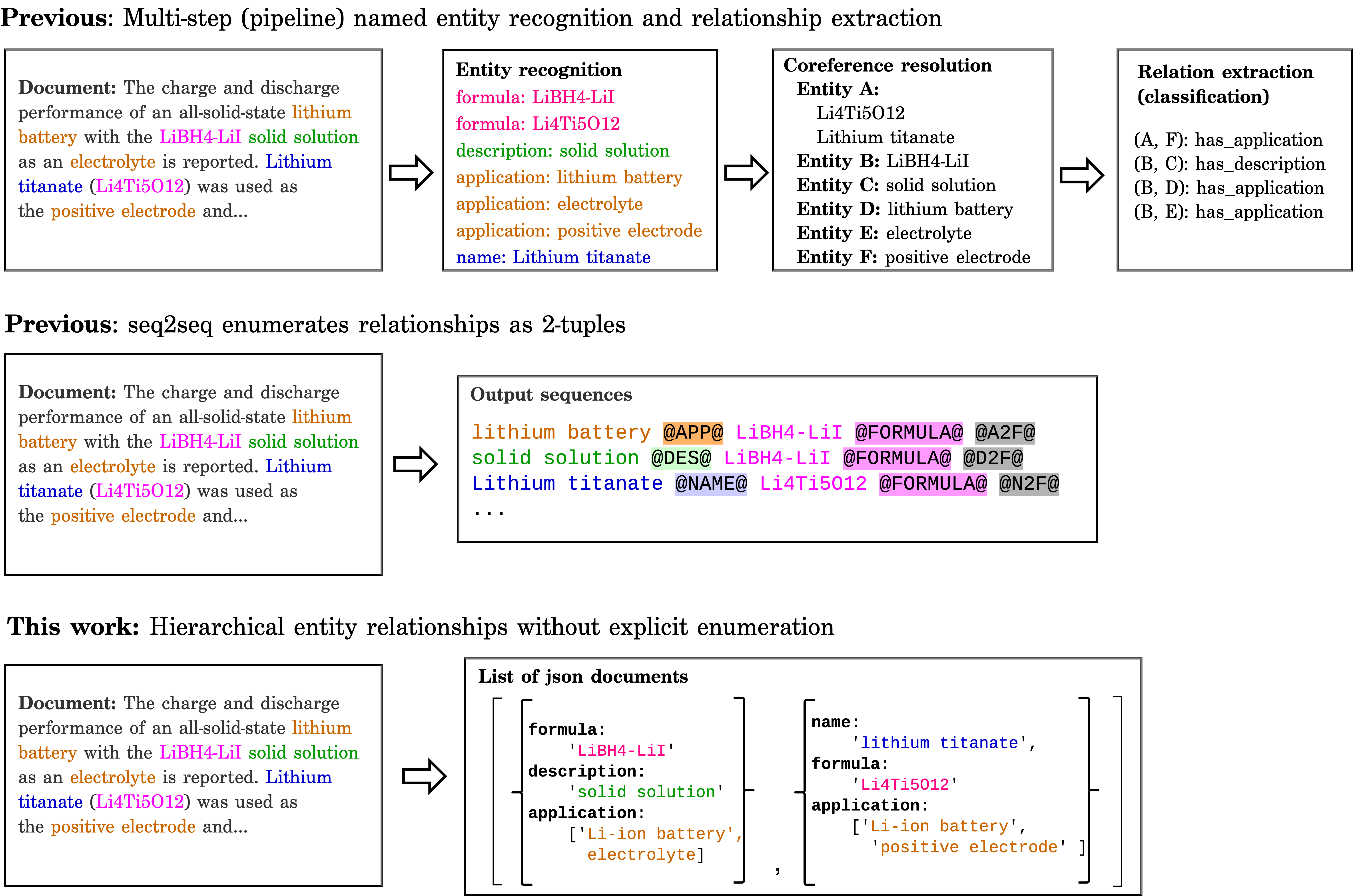}
    \caption{Simplified comparison of previous relation extraction (RE) models and the $seq2seq$-LLM approach proposed in this work.}
    \label{fig:comparison}
\end{figure*}

Yet, a key outstanding challenge in materials NLP is the development of relation extraction (RE) techniques to extract structured information that accurately describes the links \textit{between} these entities. In this paper, we describe a sequence-to-sequence approach to document-level joint named entity recognition and relation extraction (NERRE) for the extraction of complex information from scientific text (Figure \ref{fig:comparison}). The approach leverages a pre-trained large language model, GPT-3 \cite{Brown2020}, that is fine-tuned on approximately 500 document-completion examples to return structured records (\textit{e.g.,} lists of JSON documents) containing desired information. Besides its high level of accuracy, the advantages of this approach are its flexibility and accessibility. Nearly any information extraction task is accommodated by specifying a new output schema, and scientific domain experts can easily construct their own models by simply reading passages and transcribing the type of output they want the model to produce. Furthermore, publicly available language model APIs can be used in place of custom, local language models, and limited machine learning expertise is required to achieve state of the art results on a variety of complex information extraction tasks. 

\subsection*{Prior work}

Prior information extraction studies in the domain of solid-state materials include NER of chemical synthesis parameters in methods section texts \cite{Kononova2019, Huo2022, He2020, Wang2022}, quantitative results of battery cycling experiments \cite{Huang2020}, or peak absorption wavelengths for UV-Vis experiments \cite{Beard2019}, among others \cite{Weston2019, Trewartha2022, Beard2022, Kumar2022, BatteryBERT, Dong2022, Zhao2022}. Regular expressions, BiLSTM recurrent neural networks, and smaller transformer-based language models such as BERT are sufficient for such tasks. In these studies, entities (e.g., \ch{LiCoO2}, \texttt{"350K"}) rather than relations (e.g., \texttt{"350K"} is an experimental synthesis parameter for \ch{LiCoO2}) are the primary target of extraction.

Downstream tasks such as supervised machine learning or the construction of knowledge graphs require knowledge of the relationships between entities in the text. Relation extraction (RE) models are used to determine which entities are linked by a predefined set of relations. For example, in the sentence "LiCoO2 is studied as a Li-ion battery material", the material entity "LiCoO2" is linked to the application entity "Li-ion battery". There has been relatively little work on relation extraction in materials science text, but there has been much research interest in RE on general-purpose text, especially related to linking people, organizations, locations, and dates.\cite{bekoulis2018joint, Han2020} These methods have traditionally relied on pipeline-based approaches where entities are first identified by an NER model, which is followed by one or more additional steps and a final relation classification model (see Figure \ref{fig:comparison}, top row). State-of-the-art transformer-based implementations of pipeline implementations have been shown to perform document level relation extraction well on a variety of general-knowledge corpora\cite{DocRED} and more specialized domains such as chemical-disease relations\cite{Li2016} and gene-disease relations.\cite{Bravo2015}

However, scientific information often cannot be modeled as simple pairwise relations between entities. This is particularly apparent in inorganic materials science, where a compound's properties are determined by a complex combination of its elemental composition, atomic geometry, microstructure, morphology (\textit{e.g.,} nanoparticles, heterostructures, and interfaces), processing history, and environmental factors such as temperature and pressure. Furthermore, inorganic materials knowledge is often inherently hierarchical such that the relations may only be valid between one entity type and a compound entity (itself comprised of several entities and relationships). For example, we may consider a zinc oxide-based material ("ZnO" linked to a morphology "nanoparticles") to be a catalyst, but "ZnO" and "nanoparticles" alone are not necessarily catalysts in themselves. In theory, these types of relations can be modeled as $n$-tuples where $n$ is the number of entities, but comprehensively enumerating all of the possible variations is both impractical and not amenable to conventional relation extraction methods because a sufficient number of training examples is required for each relation type. Moreover, a piece of materials science knowledge may be implied for multiple simultaneous values for the same entity type. For example, a sample of an "epitaxial" "La-doped" "thin film" of HfZrO$_4$ will have different physical properties than a "La-doped" "thin film" of HfZrO$_4$ and a "La-doped" sample of HfZrO$_4$. Current RE models are not designed to practically extract or preserve such kinds of highly complex, intricately related, and hierarchical relationships between arbitrary numbers of named entities; a more flexible strategy is required.

\subsection*{Sequence-to-sequence (\textit{seq2seq}) relation extraction}

Large language models (LLMs) such as GPT-3 \cite{Brown2020}, PaLM \cite{chowdhery2022palm}, Megatron \cite{megatronturing}, OPT \cite{opt}, Gopher \cite{Hoffmann2022}, and FLAN \cite{Wei2021} have been shown to have remarkable ability to leverage semantic information between tokens in natural language sequences of varying length. They are particularly adept at sequence-to-sequence ($seq2seq$) tasks, where a text input is used to seed a text response from the model. In this paper we will refer to these inputs as "prompts" and the outputs as "completions." Use cases for seq2seq are broad\cite{bigbench} and include machine translation \cite{Dabre2020}, answering general factual knowledge questions\cite{Petroni2019, Hoffmann2022}, performing simple arithmetic \cite{Hoffmann2022}, translating between languages \cite{Dabre2020, han2022unsupervised}, summarizing text \cite{zhang2019pretraining, chowdhery2022palm}, and chatbot applications \cite{Brown2020, megatrondialogue}. It stands to reason that these models may also be also adept at complex scientific information extraction. 

Recently, end-to-end methods that use a single machine learning model have been investigated for joint named entity recognition and relation extraction (NERRE) for simple named entity recognition and pairwise relation extraction.\cite{seq2rel, rebel, Townsend2021} These methods take a sequence-to-sequence approach where a model is trained to output tuples of two or more named entities and the relation label belonging to the predefined set of possible relations between them (Figure \ref{fig:comparison}, middle row). These methods have, so far, under-performed compared to the pipeline-based approaches such as Eider\cite{Xie2021}. Fundamentally, these methods remain $n$-ary relation extraction systems and are subject to the same limitations. 

In the domain of materials science, Huang \& Cole recently fine-tuned a BERT model on battery publications and trained a model to enhance a database of NLP-extracted battery data \cite{BatteryBERT}. Their approach employed a "question and answer" (Q/A) approach that extracted limited device-level information (\textit{e.g.}, "What is the cathode?", "What is the anode?", "What is the electrolyte?") in tandem with conventional information extraction methods.\cite{BatteryBERT} We note that this approach cannot be used on passages that contain information about more than one device, and it required the BERT language model to be trained on hundreds of thousands of battery research papers before being fine-tuned on the Q/A task.

In this work, we investigate a simple and facile approach to complex information extraction where a large language model is fine-tuned for simultaneous document-level named entity recognition and relation extraction. The method is simple yet able to flexibly handle complex inter-relations (including cases where information exists in a hierarchy or as lists of multiple items) without requiring enumeration of all of possible $n$-tuple relations or preliminary NER. We fine-tune a large language model, GPT-3, to accept a text passage (for example, a research paper abstract) and write a precisely formatted "summary" of knowledge contained in the prompt. This completion can be formatted as either English sentences or a more structured schema such as a list of JSON documents. To use this method, one only has to define the desired output structure---for example, a list of JSON objects with a predefined set of keys---and annotate $\sim100-500$ text passages using this format. GPT-3 is then fine-tuned on these examples, and the resulting model is able to accurately extract desired information from text and output information in the same structured representation as shown in Figure \ref{fig:comparison}. 

This method shows strong performance on both sentence-level and document-level materials information extraction. Moreover, the method requires little knowledge of how LLMs work internally; the LLM may be simply treated by the user as a black-box that creates precisely-formatted summaries of scientific text. Therefore, researchers may use this method with little NLP experience. We also discuss how intermediate models can be used to pre-suggest entities for annotation, vastly increasing the speed and ease of annotating documents so that large training sets can be constructed relatively quickly. Although the example tasks shown are from materials science, the generality and accessibility of the method implies it may be readily applied to other domains such as physics or biology. In particular, this approach does not appear to require fine-tuning on a large corpus of domain-specific data (e.g., millions of article abstracts or paragraphs) as in previous methods; rather, the comprehensive pretraining of the LLMs along with the user-provided annotations are sufficient to accomplish a broad array of complex tasks.

\section{Methods}\label{sec:methods}

\subsection*{General seq2seq NERRE}
We fine-tune GPT-3 to perform NERRE tasks using $100-1,000$ manually annotated text-extraction (prompt-completion) pairs. Extractions contain the desired information formatted with a predefined, consistent schema across all training examples. These schemas can range in complexity from English sentences with predefined sentence structures to lists of JSON objects or nested JSON objects. In principle, many other potential schemas (\textit{e.g.,} YAML, psuedocode) may also be valid, though we do not explore those here. Once fine tuned on sufficient data adhering to the schema, a model will be capable of performing the same information extraction task on new text data with high accuracy. The model outputs completions in the same schema as the training examples. We refer to this approach generally as "LLM-NERRE".

\begin{figure*}[b!]
    \centering
    \includegraphics[width=1\textwidth]{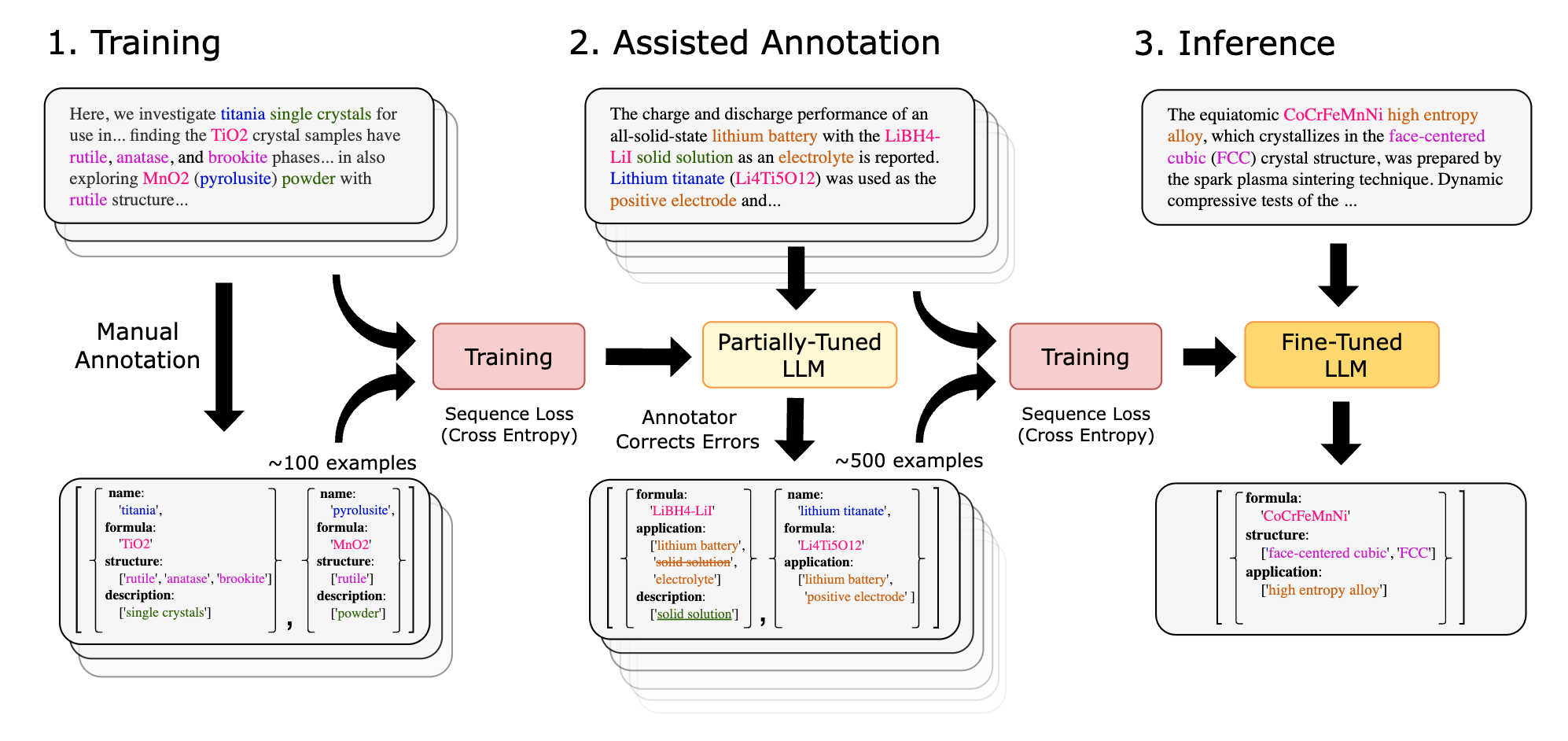}
    \caption{Overview of our sequence-to-sequence approach to document-level joint named entity recognition and relationship extraction task. In the first step, lists of JSON documents are prepared from abstracts according to a predefined schema, and a GPT-3 model is trained. In the second step, this preliminary (intermediate) model is used to accelerate the preparation of additional training data by pre-annotation with the partially trained model and manual correction. This step may be repeated multiple times with each subsequent partial fine-tuning improving in performance. In the final step, GPT-3 is fine-tuned on the complete dataset and used for inference to extract desired information from new text.}
    \label{fig:overview}
\end{figure*}

Our general workflow for training GPT-3 to perform NERRE tasks is outlined in Fig. \ref{fig:overview}. Annotations are initially performed by human domain experts to create an initial training set, and then a partially trained model is used to accelerate the collection of additional training examples. Fine-tuning is then performed on these examples to produce a "partially trained" model, which is used to pre-fill annotations that are subsequently corrected by the human annotator before being added to the training set. As discussed later in Section \ref{sec:accessibility}, we have found that this in-the-loop annotation procedure greatly accelerates annotation, reducing the average time per annotation of materials science abstracts from 100 seconds per abstract to around 40 seconds per abstract. Once a sufficient number of annotations have been completed, the final fine-tuned model is capable of extracting information in the desired format without human correction. Optionally, as illustrated in Figs. \ref{fig:example-doping}-\ref{fig:example-mofs}, the structured outputs may be further decoded and post-processed into hierarchical knowledge graphs.

\subsection*{Benchmark tasks and output schema design}
We use the described approach on three materials information extraction tasks: solid-state impurity doping, metal--organic frameworks, and general materials information extraction. Details for each are summarized in Table \ref{table:methods-parameters}. 

\begin{table*}[t!]
\centering
    \caption{Parameters for the models trained on the three materials information extraction tasks.}\label{table:methods-parameters}
  \begin{tabular}{ |c|c|c|c|c|c|c| } 
        \hline
        Task & Model Name  & Training Samples & Task level & Completion Format \\
        \hline
        \multirow{3}{*}{Doping} & \texttt{Doping-JSON}  & \multirow{3}{*}{413 sentences} & \multirow{3}{*}{Sentence} & JSON   \\\cline{2-2}\cline{5-5}
         & \texttt{Doping-ENG} & &  & \multirow{2}{*}{English sentences} \\ \cline{2-2}
         & \texttt{DopingExtra-ENG} &  & &   \\ \cline{1-2}\cline{3-3}\cline{4-5}
        MOFs & \texttt{MOF-JSON} &  507 abstracts & \multirow{2}{*}{Abstract} & \multirow{2}{*}{JSON}  \\\cline{1-2}\cline{3-3}
        General Materials & \texttt{General-JSON} &  634 abstracts &  &  \\\cline{1-2}
        \hline
  \end{tabular}
\end{table*}

\subsubsection*{Solid-state impurity doping schema}

The two core entities of interest for the solid state impurity doping task are host (\texttt{host}) and dopant (\texttt{dopant}). Hosts are defined as the host crystal, sample, or material class along with crucial descriptors in its immediate context (\textit{e.g.,} "ZnO2 nanoparticles", "LiNbO3", "half-Heuslers"). Dopants are taken to be any elements or ions that are minority species, intentionally added impurities, or specific point defects or charge carriers ("hole-doped", "S vacancies"). One host may be doped with more than one dopant (\textit{e.g.,} separate single-doping or codoping), or the same dopant may be linked to more than one host material. There may also be many independent pairs of dopant-host relations, often within a single sentence, or many unrelated dopants and hosts (no relations). We impose no restriction on the number or structure of the dopant-host relations beyond that each relation connects a host to a dopant. We also seek to extract two additional entities: "modifiers" and "result", without explicit linking (\textit{i.e.,} NER only). The \texttt{results} entity represents formulae with algebra in the stoichiometric coefficients (\textit{e.g.,} $\text{Al}_x\text{Ga}_{1-x}\text{As}$). These are formulae typically associated with crystalline solid solutions (\textit{e.g.,} $\text{CaCu}_{3-x}\text{Co}_x\text{Ti}_4\text{O}_{12}$) and their stoichiometric ranges. We also include stoichiometries where the algebra is substituted (\textit{i.e.,} $x$ value specified) and the doped result is a specific composition (\textit{e.g.,} $\text{CaCu}_{2.99}\text{Co}_{0.1}\text{Ti}_4\text{O}_{12}$). The \texttt{modifiers} is a loosely-bounded entity encapsulating other descriptors of the dopant-host relationship not captured by \texttt{dopant}, \texttt{host}, or \texttt{result}. We extract polarities (\textit{e.g.,} "n-type", "n-SnSe"), dopant quantities (\textit{e.g.,} "5 at.\%", "$x<0.3$"), defect types (\textit{e.g.,} "substitutional", "antisite", "vacancy") and other modifiers of the \texttt{host} to \texttt{dopant} relationship (\textit{e.g.,} "high-doping", "degenerately doped").  These entities (\texttt{host}, \texttt{dopant}, \texttt{result}, and \texttt{modifiers}) were chosen to define a minimal effective schema for extracting basic doping information. 

All doping-related models are trained on sentence-level prompt/completion data and produce sentence-level completions. The main motivation for this design choice is that the vast majority of dopant-related data can be found within single sentences, and the remaining relational data is often difficult to resolve consistently for both annotators and models. We expand on problems with annotations and ambiguity in the Supplementary Information. 

\begin{figure*}[htb!]
    \centering
    \includegraphics[width=1\textwidth]{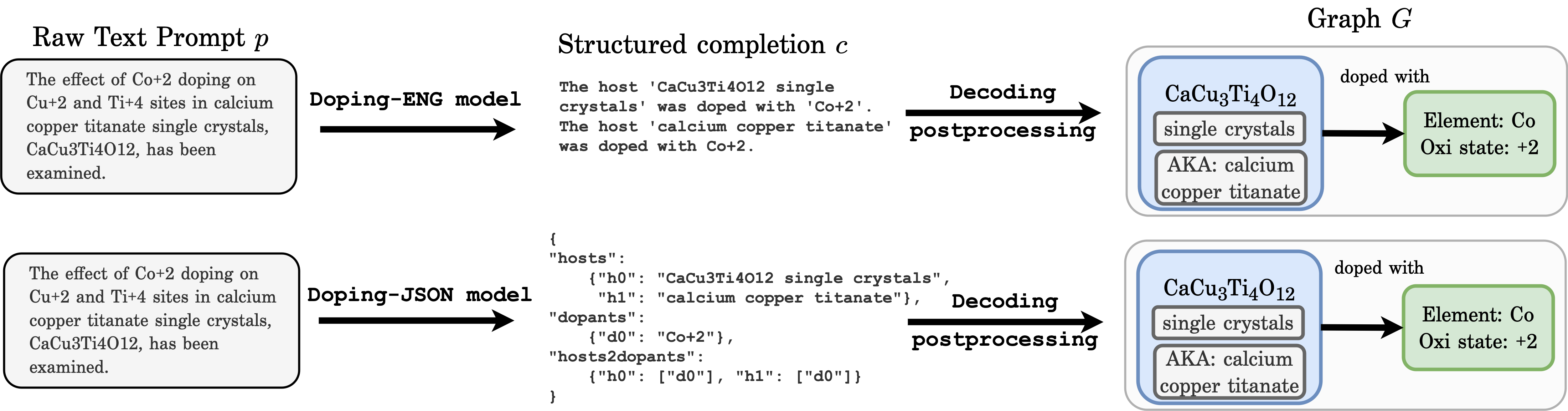}
    \caption{Annotation schema example for the doping extraction task. A raw sentence text sequence prompt $p$ is passed into an LLM-NERRE doping model which produces a structured sequence completion $c$. The structured completion format depends on the model; here, we train two separate models \texttt{Doping-ENG} and \texttt{Doping-JSON} with structured completion formats of English sentences and JSON, respectively. These models are completely independent for training and evaluation purposes. As an optional final step, the completions may be decoded and post-processed from string literals into hierarchical structured graph objects ($G$) for further analysis. Note this final step is separate from the NERRE models themselves, as the graph objects are decoded programmatically to a variety of formats (\textit{e.g.,} JSON, NetworkX objects \cite{networkx}); a more complex hierarchical graph example for the doping task is shown in Supplementary Figure \ref{fig:supp-doping-graph}.}
    \label{fig:example-doping}
\end{figure*}

We train and evaluate three separate models for the doping task. The first two, \texttt{Doping-JSON} and \texttt{Doping-ENG}, both aim to extract only \texttt{dopant} and \texttt{host} entities and the relations between \texttt{dopant} and \texttt{host}. Modifier and result entities are ignored. These models differ only in the formats of their structured completions (schemas) as shown in Fig. \ref{fig:example-doping}. The \texttt{Doping-JSON} model uses is a JSON schema representing the graph of relationships between hosts and dopants within a single sentence, where unique keys identify dopant and host strings. The model aims to learn this relatively loose schema during fine-tuning. A separate key, "hosts2dopants", describes the pairwise relations according to those unique keys. The \texttt{Doping-ENG} model encodes the entity relationships as quasi-natural language summaries. These summaries represent the same information as the JSON schema, but they more closely mimic the natural language pre-training distribution of GPT-3. This schema is filled programmatically according to the schema shown in \hyperref[sec:supp-doping-schema]{the Supplementary section on the doping schema}, where each doping entry  present is newline-separated. For example, the first line details a one-host-to-many dopants relationship, the second line expresses a dopant entity with no host, and the third line expresses a host entity with no dopant. 

Finally, we independently train and evaluate a model to retrieve \texttt{dopant} and \texttt{host} entities and relationships as well as \texttt{modifiers} and \texttt{results} entities. We label this model \texttt{DopingExtra-ENG}, as it uses the same natural-language-like schema of \texttt{Doping-ENG} but additionally extracts doping modifier and doping result data. Training data for \texttt{DopingExtra-ENG} is similar to that of the other doping models, but instead of ignoring results and modifier data, these additional entities are incorporated as additional newline-separated quasi-natural language statements. The schema for \texttt{DopingExtra-ENG} is shown in \hyperref[sec:supp-doping-schema]{the Supplementary section on the doping schema}.

\subsubsection*{General materials information schema}

\begin{figure*}[h!]
    \centering
    \includegraphics[width=1\textwidth]{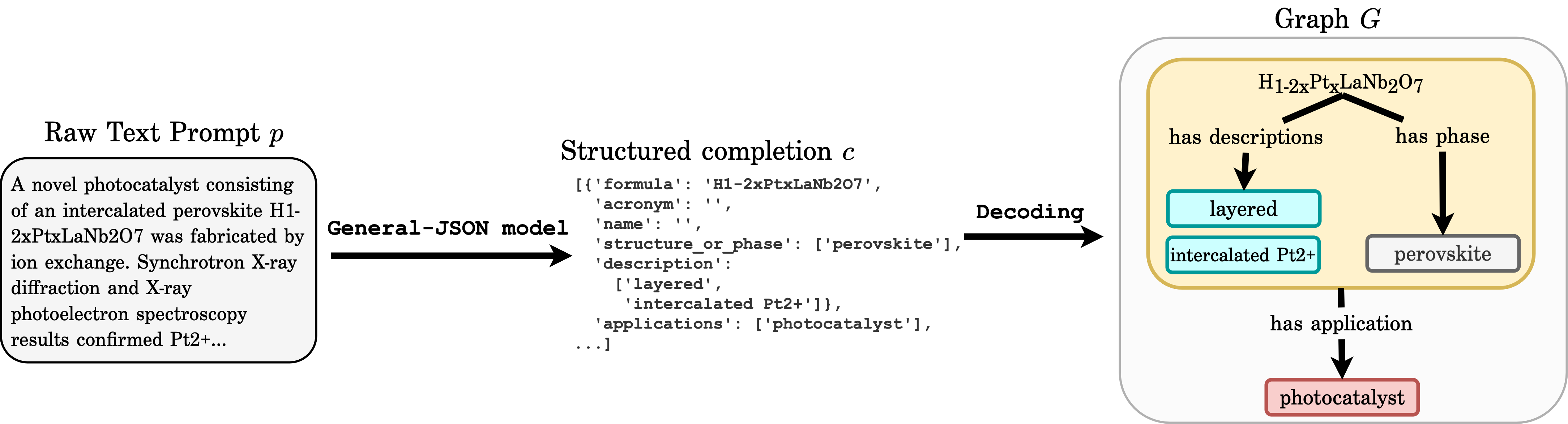}
    \caption{Annotation schema example for the general materials-chemistry extraction task. A raw full-abstract text prompt $p$ is passed to the \texttt{General-JSON} model which produces a structured completion $c$ in JSON schema. The JSON schema is a list of individual material entries ordered by appearance in the text, each of which may have a name, formula, acronym, descriptors, applications, and/or phase label. The structured completion may then be programmatically decoded to a hierarchical materials graph $G$ without a ML model.}
    \label{fig:example-general}
\end{figure*}

In our previous work \cite{Weston2019, Trewartha2022}, we focused on NER for a specific set of entity types that are particularly relevant in materials science: materials, applications, structure/phase labels, synthesis methods, \textit{etc.} However, we were not able to link these labeled entities together to record their relations beyond a simple "bag-of-entities" approach. In this work, we train an LLM to perform a "general materials information extraction" task that captures both entities and the complex network of interactions between them. 

The schema we have designed for this task encapsulates an important subset of information about solid compounds and their applications. Each entry in the list, a self-contained JSON document, corresponds one-to-one with a material mentioned in the text. Materials entries are ordered by appearance in the text. The root of each entry starts with a compound's name and/or its chemical formula. If a name or formula is not mentioned for a material, no information about that material is extracted from the text. We also extract acronyms mentioned for a material's name/formula, although in cases where only an acronym is mentioned we do not create a material entry for the compound. Compounds that are not solids (ions, liquids, solvents, solutions, etc) are generally not extracted. The \texttt{name}, \texttt{formula}, and \texttt{acronym} fields are exclusively given string value in the JSON document for each material whereas the \texttt{description}, \texttt{structure\_or\_phase}, and \texttt{applications} fields are lists of an arbitrary number of strings. We label this model \texttt{General-JSON}, and an example is shown in Fig. \ref{fig:example-general}.

Descriptions are defined as details about a compound's processing history, defects, modifications, or the sample's morphology. For example, consider the hypothetical text "Pt supported on CeO2 nanoparticles infused with Nb...". In this case, the "descriptions" entry for the material document referring to "Pt" might be annotated as \texttt{"['supported on CeO2']"}, and the entry in the document referring to "CeO2" would be \texttt{"['nanoparticles', 'Nb-doped']"}.

Structures/phases are defined as entities that directly imply the crystalline structure or symmetry of the compound. Crystal systems such as "cubic" or "tetragonal", structure names such as "rutile" or "NASICON", and space groups such as "Fd3m" or "space group No. 93" are all extracted in this field. We also include any information about crystal unit cells, such as lattice parameters and the angles between lattice vectors. "Amorphous" is also a valid structure/phase label.

Applications are defined as high-level use cases or major property classes for the material. Applications may represent different levels of device-level implementation. For example, a battery cathode material may have \texttt{"['Li-ion battery', 'cathode']"} as its applications entry. Generally, applications are mentioned in the order they are presented in the text, except for certain cases such as battery materials, in which case the type of device is generally mentioned before the electrode type, and catalysts, where the reaction catalyzed is generally listed following the "catalyst" entity in the list (\textit{e.g.,} \texttt{"['catalyst', 'hydrogenation of citral']"}). More details and caveats of this schema are given in the \hyperref[sec:supp]{Supplementary Information}.

\subsubsection*{Metal--organic framework (MOF) schema}

\begin{figure*}[h!]
    \centering
    \includegraphics[width=1\textwidth]{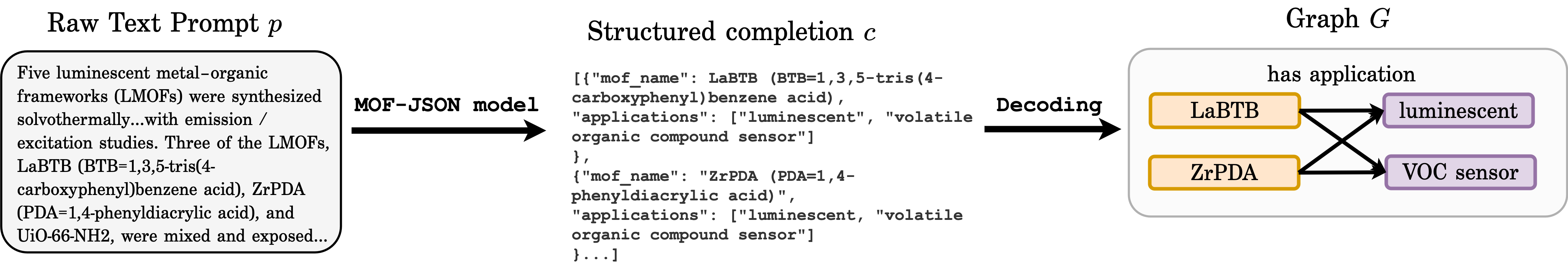}
    \caption{Annotation schema example for the metal--organic frameworks extraction task. A raw full-abstract text prompt $p$ is fed into the \texttt{MOF-JSON} model which produces a structured output sequence $c$ similar to that of the \texttt{General-JSON} model shown in Fig. \ref{fig:example-general}. The output sequence is a formatted string literal which can be directly loaded as JSON. The string may then be optionally decoded to a graph $G$ for further analysis. In this example, only the MOF name and application were extracted from the passage, and both MOFs (LaBTB and ZrPDA) are linked to both applications (luminescent and VOC sensor).}
    \label{fig:example-mofs}
\end{figure*}

The schema used for the MOF cataloging task is based on the general materials information schema described in the previous section, which was modified to better suit the needs of MOF researchers. We developed this schema to extract MOF names (\texttt{name}), an entity for which there is no widely accepted standard\cite{bucior2019identification}, and chemical formulae (\texttt{mof\_formula}), which form the root of the document. If no name or formula is present, no information is extracted for that instance. In addition, because there is a great deal of interest in using MOFs for ion and gas separation\cite{li2021metal,qian2020mof}, we extract "guest species", which are chemical species that have been incorporated, stored, or absorbed in the MOF. We extract applications the MOF is being studied for as a list of strings (e.g. \texttt{"['gas-separation']"} or \texttt{"[heterogeneous catalyst', 'Diels-Alder reactions']"}) as well as relevant descriptors for the MOF/sample, such as its morphology or processing history, similar to the general information extraction schema. Entries in the list are generally added in the order the material names/formulas appear in the text (described in more detail in Supplementary Information). The MOF extraction model is labeled \texttt{MOF-JSON}, and an example is shown in Fig. \ref{fig:example-mofs}.

\subsection*{Comparison benchmarks and evaluation}

To compare our model with other sequence-to-sequence approaches to information extraction, we perform a benchmark of two methods on the doping task to compare to the LLM-NERRE models. The first employs the seq2rel method of Giorgi \textit{et al.}\cite{seq2rel} for the host-dopant task. We formatted host-dopant relationships under tags labeled \texttt{@DOPANT@} and \texttt{@BASEMAT@} (base/host material), with their relationship signified by \texttt{@DBR@} ("dopant-base material relationship"); these sequences were constructed from the same training data as the \texttt{Doping-JSON} and \texttt{Doping-ENG} models. We trained seq2rel to perform sentence-level extraction with 30 epochs, batch size of 4, encoder learning rate $2\cdot10^{-5}$, decoder learning rate $5\cdot10^{-4}$, and pretrained BiomedNLP BERT tokenizer\cite{pubmedbert} (further training details are in \hyperref[supp:seq2rel-params]{Supporting Information}). Additionally, we compare against the previously published MatBERT doping-NER model \cite{Trewartha2022} combined with proximity-based heuristics for linking. With this method, a MatBERT NER model pretrained on $\sim50$ million materials science paragraphs and fine-tuned on 455 separate manually annotated abstracts first extracts hosts and dopants and then links them if they co-occur in the same sentence.


\subsection*{Datasets}

Datasets were prepared from our database of more than 5 million research paper abstracts.\cite{Tshitoyan2019} Annotations were performed by human annotators using a graphical user interface built using Jupyter\cite{Kluyver2016jupyter}, although in principle annotations could be conducted via a simple text editor. To accelerate the collection of training data, new annotations are collected via a "human in the loop" approach where models are trained on small datasets and their outputs are used as starting points and corrected by human annotators (see Figure \ref{fig:overview}.) This process of training and annotation is completed multiple times until a sufficiently large set of training data was achieved. Single domain experts separately annotated each of the training and test sets.

\subsubsection*{Doping dataset}
Training and evaluation data was gathered from our database of research paper abstracts using the keywords "n-type", "p-type", "-dop", "-codop", "doped", "doping", and "dopant" (with exclusions for common irrelevant keywords such as "-dopamine"), resulting in $\sim375\text{k}$ total abstracts. All doping tasks were trained on text from 162 randomly selected abstracts, comprising 1,215 total sentences and filtered with regular expressions to only include 413 relevant (potentially including doping information) sentences. Doping tasks were tested on an additional 232 sentences (77 relevant by regex) from a separate holdout test set of 31 abstracts.

\subsubsection*{General materials dataset}
Training and evaluation data was gathered from our abstract database using keywords for a variety of materials properties and applications (\textit{e.g.,} "magnetic", "laser", "space group", "ceramic", "fuel cell", "electrolytic"). A domain specialist used \textit{ad hoc} class balancing to select keywords resulting in $\sim10-50$ abstracts per-keyword; all abstracts were manually screened for relevance. More details on this data collection are provided in \hyperref[sec:supp]{Supplementary Information}. The resulting $\approx$650 entries were manually annotated with the general materials information schema. Results were evaluated using a 10\% random sample for validation, and this procedure was averaged over five trials with no hyperparameter tuning. 

\subsubsection*{Metal--organic framework dataset}
Training and evaluation data was selected from our database using the keywords ”MOF”, ”MOFs”, ”metal-organic framework”, ”metal organic framework”, ”ZIF”, ”ZIFs”, ”porous coordination polymer”, and ”framework material”, which produced approximately 6,000 results with a very high likelihood of containing MOF-related information. From these, 507 abstracts were randomly selected and annotated by a human MOF expert. Results were evaluated using the same procedure as the general materials dataset in the previous section.

\subsection*{GPT-3 fine tuning details}
For all tasks, we fine-tune GPT-3 (`davinci', 175B parameters)\cite{Brown2020} using the OpenAI API, which optimizes the cross-entropy loss on predicted tokens. Doping models were trained for 7 epochs at a batch size of 1, with inference temperature of 0 and output limited to a maximum length of 256 tokens (all doping models) or 512 tokens (\texttt{General-JSON}, \texttt{MOF-JSON}). The intermediate models shown in Figure \ref{fig:learning-curve} were trained with a number of epochs depending on the number of training samples $t$: 2 epochs for $2^0 \leq t < 2^6$, 4 epochs for $2^6 \leq t \leq 2^7$, and 7 epochs for $t = 2^8$. Models for the MOF and general materials extraction tasks were trained for 4 epochs with a batch size of 1. We use a learning rate multiplier of 0.1 and a prompt loss weight of 0.01 but have not performed hyperparameter tuning for these hyperparameters. For all tasks, the start and end tokens used were \texttt{"\textbackslash n\textbackslash n\textbackslash n\#\#\#\textbackslash n\textbackslash n\textbackslash n"} and \texttt{"\textbackslash n\textbackslash n\textbackslash nEND\textbackslash n\textbackslash n\textbackslash n}.

\subsection*{Evaluation criteria}

The fuzzy and complex nature of the entities and relationships detailed in the previous section necessitates the use of several metrics for scoring. We evaluate the performance of all models on two levels: sequence reconstruction and information extraction performance. 

\subsubsection*{Sequence reconstruction}

We measure sequence reconstruction as the ability of the model to output a sequence which is parsable (\textit{i.e.,} output adheres to the schema format) and is similar to the true output sequence. In this evaluation, we do not decode entries or inspect individual entities or relationships; we merely measure the ability of the model to output a string similar to the true string.

We report three metrics: exact sequence match accuracy, Jaro-Winkler similarity \cite{Winkler1999}, and parsability. Each of these metrics, averaged over all samples to evaluate for each task, is defined on $[0, 1]$ where 0 indicates no sequence reconstruction and 1 indicates perfect reconstruction. Details and formalism on each metric are given in the Supplementary Info.

\subsubsection*{Information extraction performance}

We measure information extraction performance as the ability of the model to jointly recognize entities and the relationships between them. We report two methods of scoring entity-relation data:

\begin{enumerate}
    \item A triplet $F_1$ (defined later) computed on a stringent exact word-match basis (\textit{i.e.,} how many words are correctly linked together exactly as they appear in the source text prompt).
    \item A less-stringent triplet $F_1$ based on manual inspection from a domain expert (\textit{i.e.,} how many sets of words are approximately correct in the context).
\end{enumerate}

\textbf{Exact word-match basis scoring.} We score named entity relationships on a word-basis by first converting an entity $E$ into a set of constituent $k$ whitespace-separated words $E = \{w_1, w_2, w_3, ..., w_k\}$. Comparing two entities
$E^{\text{true}}$ and $E^{\text{test}}$ for NER only, we count the number of exactly matching words in both sets as true positives ($E^{\text{true}} \cap E^{\text{test}}$) and the mathematical differences between the sets as false positives ($E^{\text{test}} - E^{\text{true}}$) or false negatives ($E^{\text{true}} - E^{\text{test}}$). For example, if the true entity is "Bi2Te3 thin film" and the predicted entity is "Bi2Te3 film sample", we record two true positive word exact matches ("Bi2Te3", "film"), one false negative ("thin"), and one false positive ("sample"). The lone exception to this method is for formula-type entities crucial for identifying materials, in which case $E^{\text{test}}$ must contain \textit{all} $w_i$ parsable as stoichiometries for \textit{any} of $w_i \in E^{\text{test}}$ to be considered correct.  For example, if the true entity is "Bi2Te3 thin film", and the predicted entity is "thin film", we record three false negatives. Thus, any formula-type entity (Doping \texttt{host}, Doping \texttt{dopant}, General \texttt{formula}, and MOF \texttt{mof\_formula}) containing a chemical composition is entirely incorrect if the composition is not an exact match. This choice of evaluation was made to avoid metrics artificially inflating the performance of the model. For an example, if we evaluate a true "Bi2Te3 nanoparticles" to a predicted "Bi2Se3 nanoparticles", we obtain very high similarities via Jaro-Winkler (0.977) and character-level BLEU-4 (0.858). In reality, this example prediction should be recorded as entirely incorrect -- the material's chemistry is wrong. 

We now score relationships between entities on a word basis using on the number of correct triplets. Triplets are 3-tuples relating word $w^n_j$ of an entity $E_n$ to word $w^m_k$ of an entity $E_m$ by relationship $r$, represented as $(w^n_j, w^m_k, r)$. The total set of correct relationships $T^{\text{true}}$ for a text contains many of these triplets. A test set of relationships $T^{\text{test}}$ is evaluated by computing the number of triplets found in both sets ($T^{\text{true}} \cap T^{\text{test}}$) as true positives and the differences between these sets as false positives ($T^{\text{test}} - T^{\text{true}}$) or false negatives ($T^{\text{true}} - T^{\text{test}}$). Entity triplets are also bound to the same requirement for composition correctness if either of the words in the triplet belong to an formula-type entity (\texttt{host}, \texttt{dopant}, \texttt{formula}, \texttt{mof\_formula}), \textit{i.e.,} we count all triplets for two entities as incorrect if the formula is not an exact string match. With correct and incorrect triplets identified, $F_1$ scores for each relation are calculated as:

\begin{equation}
    \text{precision} = \frac{\text{No. of correct relations retrieved}}{\text{No. of relations retrieved}}
\end{equation}

\begin{equation}
    \text{recall} = \frac{\text{No. of correct relations retrieved}}{\text{No. of relations in test set}}
\end{equation}

\begin{equation}
    F_1 = \frac{2(\text{precision} \cdot \text{recall})}{\text{precision} + \text{recall}}
\end{equation}

To compute triplet scores across entire test sets in practice, we first select a subset of relations to evaluate. We note that this is not a full evaluation of the task we are training the model to perform, which involves linking many interrelated entities \textit{simultaneously}, but is rather provided to help give a general sense of its performance compared to previous NERRE methods. For the doping task, we evaluate \texttt{host}-\texttt{dopant} relationships. For the general materials and MOF tasks, we evaluate relationships between the formula field (\texttt{formula} for general materials, \texttt{mof\_formula} for MOFs) and all other remaining fields. For \texttt{description}, \texttt{structure\_or\_phase}, and \texttt{applications} fields, all of which may contain multiple values, all of the possible formula-value pairs are evaluated.


\textbf{Manual evaluation.}\label{sec:manual-evaluation} The metrics provided in prior sections demonstrate automatic and relatively strict methods for scoring NERRE tasks, but the underlying capabilities of the LLM models are best shown with manual evaluation. This is most apparent in the case of the \texttt{General-JSON} model, where exact boundaries on entities are fuzzier, precise definitions are difficult to universally define, and annotations include some implicit entity normalization. For example, the text "Pd ions were intercalated into mesoporous silica" may have equivalently have a correct \texttt{description} field for the material "silica" including "Pd-intercalated", "Pd ion-intercalated", "intercalated with Pd ions", \textit{etc.}; the exact choice of which particular string is used as the "correct" answer is arbitrary.

To better address scoring of these fuzzy tasks, we introduce an adjusted score based on a domain expert's manual evaluation of whether the information extracted is a valid representation of the information actually contained in the passage. We term this adjusted score "manual information extraction score"; it constitutes a basis for precision, recall, and $F_1$ that quantifies the quality of overall information capture for cases where there may be equivalent or multiple ways of representing the same concept. This score was constructed to better estimate the performance of our model for practical materials information extraction tasks.

We score entities extracted by annotators but not present in the model's output as false negatives, except when reasonable variations are present. The criteria for a true positive are as follows:
\begin{enumerate}
    \item The entity comes from the original passage or is a reasonable variation of the entity in the passage (\textit{e.g.,} "silicon" $\longrightarrow$ "\ch{Si}"). It is not invented by the model.
    \item The entity is a root entity or is grouped with a valid root entity. For the \texttt{General-JSON} model, a root entity is either a material's formula or name. If both are present, the formula is used at the root.
    \item The entity is in the correct field in the correct root entity's group (JSON object). 
\end{enumerate}

Manual information extraction scores are reported per-entity as if they were NER scores. However, the requirements for a true positive implicitly include relational information, since an entity is only correct if is \textit{grouped with} the correct root entity. 




\section{Results}\label{sec:results}

We first report results of the sequence-level evaluation metrics as described in Section \ref{sec:methods} in order to give the reader an idea of how likely an output sequence from an LLM-NERRE model is to be exactly correct, somewhat correct (via Jaro-Winkler), or parsable (\textit{i.e.,} well-formatted and compliant with the schema the model was trained on). Sequence-level metrics represent at a high-level how likely the model is to "follow the rules". Next, we evaluate models on the materials NERRE tasks with more granularity for individual relations and entities, and we compare with a BERT-based proximity approach and seq2rel; these results will allow the reader to obtain a more in-depth view of model-performance. Finally, we use the \texttt{General-JSON} model to compare against BatteryBERT, and examine the effect of training set size on performance for the doping task. 

\subsection*{Sequence-level results}
\begin{figure}
    \centering
    \includegraphics[width=.95\columnwidth]{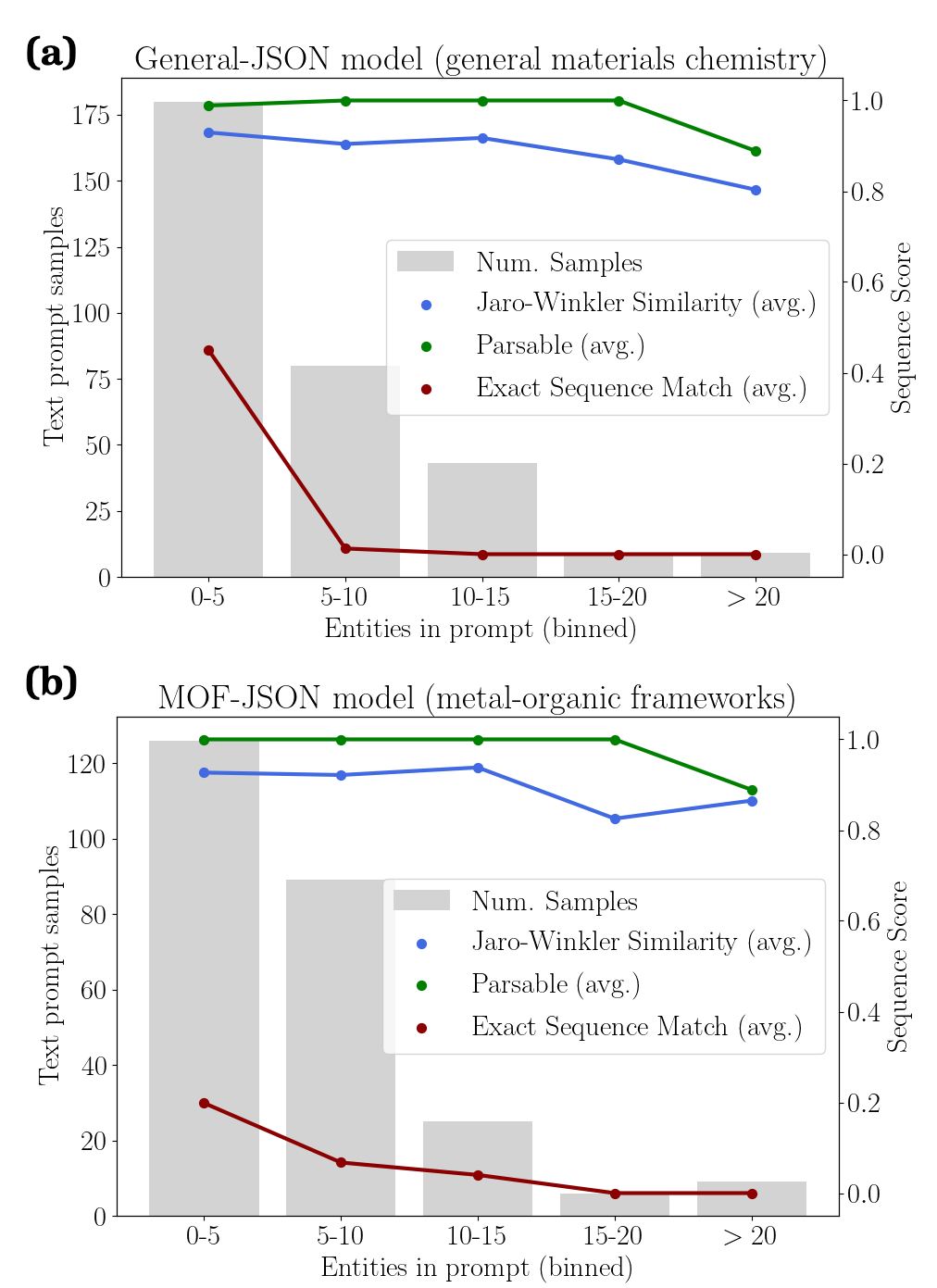}
    \caption{Sequence reconstruction metrics segmented by number of entities in the prompt for \textbf{(a)} The \texttt{General-JSON} model and \textbf{(b)} the \texttt{MOF-JSON} model. The $x$ axis defines the bins by the number of true entities in the test-set sample; as we progress from left to right, the bins represent more complex samples. The left $y$ axis (corresponding to the grey bars) shows the number of samples in each bin. The right $y$ axis (corresponding to the colored lines centered on each bin) shows the average scores of each bin by the metrics of exact string matching (red), Jaro-Winkler similarity (blue), and parsability (green). While exact string matches are very unlikely responses for the models for highly complex samples (20+ entities), the models are able to retain high sequence similarity and parsability.}
    \label{fig:sequence-scores-per-bin}
\end{figure}

Table \ref{table:results-sequence-accuracy} shows results for sequence-level matching. We find a wide variability between models for reconstructing test set output sequences exactly, with the simpler doping models containing exact matches on $58.4-64.9\%$ of test sequences. The more complex tasks, \texttt{MOF-JSON} and \texttt{General-JSON}, have exact match accuracies ranging from $12.5$ to $25.6\%$ of test sequences. Exact matches are inherently less probable with longer output sequences, as even a single error in a very long output sequence results in an exact match failure. This trend is easier to see with the test samples' sequence reconstruction scores binned into groups based on the number of entities each test set true sample contains. As shown in Fig. \ref{fig:sequence-scores-per-bin} for \texttt{MOF-JSON} and \texttt{General-JSON}, exact match sequence reconstruction is the highest for the simplest entries (\textit{i.e.,} those with the lowest number of entities). 

All models have average Jaro-Winkler similarities $\geq 91.7\%$ and parsabilities $\geq 98.7\%$. In contrast to the exact match score, Jaro-Winkler similarity and parsability do not degrade as rapidly with increasing complexity (for which we use true number of entities as a proxy) of the prompt. This is encouraging, as the model is able to retain consistent, parsable formatting even in very long output sequences with many (20+) interrelated and hierarchical entities. 

\begin{table*}[!tb]
\centering
    \caption{Sequence-level error metrics for completions for all tasks, evaluated and averaged over the test set for each task. Completions are only considered correct in the exact match if the full output sequence $c$ is recovered exactly. Average Jaro-Winkler similarities are also shown to measure string similarity continuously on $[0,1]$, where $1$ indicates a perfect match and $0$ indicates no match. Parsability, the ability to format $c$ in the same manner as the training schema, is recorded for each sample as 0 (not parsable) or 1 (parsable).  The average exact correctness, Jaro-Winkler similarity, and parsability are normalized to the range $0-100\%$.}\label{table:results-sequence-accuracy}
    
    \centering
  \begin{tabular}{ |c|c|c|c|c| } 
        \hline
        \multirow{2}{*}{Task} & \multirow{2}{*}{Model Name}  & Exact Match  & Avg. Jaro-Winkler  & Parsable and \\
        & & Sequence Accuracy (\%) & Similarity (\%) & Decodable (\%) \\
        \hline
        \multirow{3}{*}{Doping} & \texttt{Doping-ENG}  & 59.7 & 94.3 & 98.7 \\\cline{2-5}
            & \texttt{Doping-JSON} & 64.9 & 96.7 & 100  \\\cline{2-5}
            & \texttt{DopingExtra-ENG} & 58.4 & 93.8 & 98.7 \\
        \hline
        MOFs & \texttt{MOF-JSON} & 12.5 & 92.1 & 99.7 \\
        \hline
        General materials & \texttt{General-JSON} & 25.6 & 91.7 & 99.1\\
        \hline
  \end{tabular}
  \vspace{5px}
\end{table*}

\subsection*{Relation extraction performance}\label{sec:perform-erm}

In Tables \ref{table:results-linking-doping} to \ref{table:results-linking-mofs} we report the NERRE scores computed using the exact word-match basis detailed in Section \ref{sec:methods}. We note that the pure NER and RE scores are somewhat intertwined because a model must correctly recognize entities in order to correctly extract a relationship between them. The scores in Tables \ref{table:results-linking-doping}-\ref{table:results-linking-mofs} thus reflect the entire NERRE (NER + RE) task, not NER or RE separately. The raw NER-only scores broken down by entity for each of the tasks are shown in the Supplement in Table \ref{table:results-recognition}.

\subsubsection*{Doping task}
 We find the LLM-NERRE models have significantly higher $F_1$ than either the MatBERT-Doping + Proximity or seq2rel models (Table \ref{table:results-linking-doping}), though the LLM-NERRE models are not superior across all metrics (\textit{e.g.,} recall). As an example, consider the MatBERT-Doping + Proximity model, a model which represents the performance of a state-of-the-art fine-tuned NER model coupled with a simple proximity co-occurrence rule for linking dopants to hosts. While this model is able to extract entity relationships from text with comparable recall (0.714) to the LLM-NERRE models, its overall $F_1$ is reduced by a comparatively poor precision (0.441). As shown in Supplement Table \ref{table:results-recognition}, this is in part caused by low NER precision for \texttt{host} entities. The discrepancy between the MatBERT-Doping model's NER performance here and the NER performance in the original publication (NER dopant $F_1 = 0.82\pm0.02$ and basemat $F_1 = 0.72 \pm 0.02$) \cite{Trewartha2022} may be explained by our test samples being more qualitatively complex. The remaining loss in MatBERT-Doping + Proximity $F_1$ is due to the error inherent in the simple proximity rule defined for linking; essentially, not every host material is necessarily related to every dopant specie within a single sentence. We find higher precision (0.650) and recall (0.550) from the seq2rel model resulting in $F_1 = 0.596$ slightly higher than MatBERT-Doping + Proximity but substantially lower than any of the LLM-NERRE models. We acknowledge that the initial seq2rel model was derived from the PubMedBERT \cite{pubmedbert} pretrained BERT model as per the original implementation \cite{seq2rel}, and the seq2rel method may be improved by using a BERT model pretrained exclusively on materials text rather than biomedical text.
 
 Notably, all three LLM-NERRE models exceed the performance of the two baselines in both NERRE and pure NER performance, regardless of schema, despite being trained on a much smaller training dataset than the MatBERT-NER model (413 sentences vs. 455 abstracts). We find $F_1 > 0.7$ across all doping models, with the best performance from the \texttt{DopingExtra-ENG} model. The higher performance for this model compared to the less complex \texttt{Doping-ENG} model may be explained by the higher entity recognition score on the \texttt{dopant} entity. The high NERRE recall, precision, and $F_1$ for \texttt{DopingExtra-ENG} is particularly interesting because \texttt{DopingExtra-ENG} is both more accurate and more capable (\textit{i.e.,} extracts results and modifiers entities in addition to host-dopant relationships) than \texttt{Doping-ENG} and \texttt{Doping-JSON}, despite being trained on the same number of samples. This may be due to the model learning during training that some result/modifier entities (otherwise unlabelled in the \texttt{Doping-ENG} and \texttt{Doping-JSON models}) are not hosts or dopants, and the model predicts fewer false positive hosts or dopants. The \texttt{DopingExtra-ENG} model therefore is a notable example of asking an LLM-NERRE model to "do more" while having no significant drawback in $F_1$. 

    \begin{table*}[!htb]
    \caption{Exact match NERRE scores for the (host, dopant) linking task, evaluated on a per-word basis. Links are only correct if both entities and the relationship are correct. The highest score in each category are shown in bold. Note the \texttt{DopingExtra-ENG} scores here refer to the same task as the other doping models, excluding metrics for extra information (\texttt{result}, \texttt{modifier}). }\label{table:results-linking-doping}
    \centering
  \begin{tabular}{ |c|c|c|c|c| } 
    \Xhline{2\arrayrulewidth}
        Method & Recall (Exact Match)  & Precision (Exact Match)  & $F_1$ (Exact Match)  \\\Xhline{2\arrayrulewidth}
        MatBERT-Doping\cite{Trewartha2022} + Proximity & 0.714 & 0.441 & 0.545 \\\hline
        seq2rel & 0.65 & 0.55 & 0.60 \\\hline
        \texttt{Doping-ENG} (this work) & 0.776 & 0.789 & 0.783\\\hline
        \texttt{Doping-JSON} (this work) &  0.684 & 0.758 & 0.719 \\\hline
        \texttt{DopingExtra-ENG} (this work) & \textbf{0.828} & \textbf{0.872} & \textbf{0.849} \\\hline
    \Xhline{2\arrayrulewidth}
  \end{tabular}
    \vspace{15px}
\end{table*}

\subsubsection*{General and MOF tasks}

As shown in Tables \ref{table:results-linking-general} and \ref{table:results-linking-mofs}, NERRE exact word-match $F_1$ scores for the \texttt{General-JSON} and \texttt{MOF-JSON} models are generally lower than those of the doping models. We reiterate to the reader that these two tasks are both vastly more complex than the doping task; rather than focusing on a single essential relation (host, dopant) and its entities, these models must simultaneously extract many entity types and potential relations as groups. For example, the \texttt{General-JSON} model must simultaneously extract acronyms, names, formulae, phase labels, device applications, and material descriptions and determine all relationships between all entities. For simplicity, we report only the most essential of these relationships in the exact word-match scores, namely for links between material formula (for general) or MOF name (for MOFs) and all other entities in each model's schema. 
For the \texttt{General-JSON} model, the highest NERRE word-match score ($F_1 = 0.613$) was found for the material \texttt{formula} $\longrightarrow$ \texttt{name} relationship while the lowest ($F_1 = 0.341$) was found for \texttt{formula} $\longrightarrow$ \texttt{description}. The relatively lower fidelity of formula links to applications, structures, and descriptions may be caused by the generally fuzzier boundaries around these entities in the annotations, particularly in the case of description entities. For MOFs, links between guest species and specifically named MOFs are recovered most reliably ($F_1 = 0.606$) while links between MOF names and descriptions are the least reliable. This is again likely due to how reliably the expert annotator is able to delimit these entities, as guest species (\textit{e.g.,} "Hg2+") are qualitatively easier to delimit than descriptions (\textit{e.g.,} "mesostructured MOFs formed by Cu2+ and 5-hydroxy-1,3-benzenedicarboxylic acid").  

    \begin{table*}[!htb]
    \caption{Scores for the \texttt{General-JSON} LLM-NERRE model on the general materials information extraction task, evaluated on an exact word-match basis. Links are only correct if both entities and the relationship are correct. 
    }\label{table:results-linking-general}
    \centering
  \begin{tabular}{ |c|c|c|c|c| } 
    \Xhline{2\arrayrulewidth}
        Relation & Recall (Exact Match)  & Precision (Exact Match)  & $F_1$ (Exact Match)  \\\Xhline{2\arrayrulewidth}
        (material formula, material name) & 0.539 & 0.716 & 0.613 \\\hline
        (material formula, acronym) & 0.470 & 0.635 & 0.537 \\\hline
        (material formula, application) & 0.470 & 0.499 & 0.481 \\\hline
        (material formula, structure) & 0.368 & 0.411 & 0.388 \\\hline
        (material formula, description) & 0.304 & 0.392 & 0.341 \\\hline
    \Xhline{2\arrayrulewidth}
  \end{tabular}
    \vspace{15px}
\end{table*}

    \begin{table*}[!htb]
    \caption{Scores for the \texttt{MOF-JSON} model on the the metal--organic framework (MOF) information extraction task, evaluated on an exact word-match basis. Links are only correct if both entities and the relationship are correct. 
    }\label{table:results-linking-mofs}
    \centering
  \begin{tabular}{ |c|c|c|c|c| } 
    \Xhline{2\arrayrulewidth}
        Relation & Recall (Exact Match)  & Precision (Exact Match)  & $F_1$ (Exact Match)  \\\Xhline{2\arrayrulewidth}
        (MOF name, formula) & 0.409 & 0.455 & 0.424 \\\hline
        (MOF name, application) & 0.461 & 0.560 & 0.504 \\\hline
        (MOF name, guest species) & 0.588 & 0.665 & 0.606 \\\hline
        (MOF name, description) & 0.247 & 0.464 & 0.318 \\\hline
    \Xhline{2\arrayrulewidth}
  \end{tabular}
    \vspace{15px}
\end{table*}

However, these word-matching scores imply LLM-NERRE models perform worse than what is actually observed. As described in Section \ref{sec:methods}, we also report manually evaluated information extraction scores (Table \ref{table:manually-computed-f1}) for the general task for a 10\% random subset of the total test set. These manually-evaluated scores are more representative of the true (end result, human observed) performance of the LLM-NERRE models because they account for ambiguity in how the same information can be written (e.g. "Au nanoparticles" vs "gold nanoparticles"). When manually examined by a domain expert, the information extraction $F_1$ scores increase dramatically for formulae (0.943 vs 0.537), materials names (0.848 vs 0.613), applications (0.832 vs 0.481), structures (0.829 vs 0.388), and descriptions (0.704 vs 0.341). 

We find acronyms have the lowest manual information extraction scores, which we attribute to the fact that acronyms are relatively rare in the training class compared to the others (appearing in only 52 abstracts across the entire dataset, $\sim9\%$ of the documents) and that material acronyms often can be confused with chemical formulas (e.g. "AuNP" is the acronym for gold nanoparticle but is also a valid chemical formula). Usually context clues are the only way to disambiguate cases like this, and including more training data with acronyms may improve the acronym extraction score further.

    \begin{table*}[!htb]
    \caption{Manual information extraction scores for the general materials task. Scores measure the model's ability to extract inter-related data together (i.e. assigning entities correct labels and grouping them appropriately, as described in Section \ref{sec:manual-evaluation}.) }\label{table:manually-computed-f1}
\centering
  \begin{tabular}{ |c|c|c|c| } 
        \Xhline{2\arrayrulewidth}
        Entity & Extraction Recall & Extraction Precision & Extraction $F_1$ \\
        \Xhline{2\arrayrulewidth}
        name & 0.692 & 1.0 & 0.818 \\\hline
        formula & 0.943 & 0.943 & 0.943 \\\hline
        acronym & 0.667 & 0.400 & 0.500 \\\hline
        applications & 0.797 & 0.870 & 0.832 \\\hline
        structure or phase & 0.754 & 0.920 & 0.829 \\\hline
        description & 0.576 & 0.905 & 0.704 \\\hline
  \end{tabular}
\end{table*}

\subsection*{Effect of training set size}

\begin{figure}
    \centering
    \includegraphics[width=.8\columnwidth]{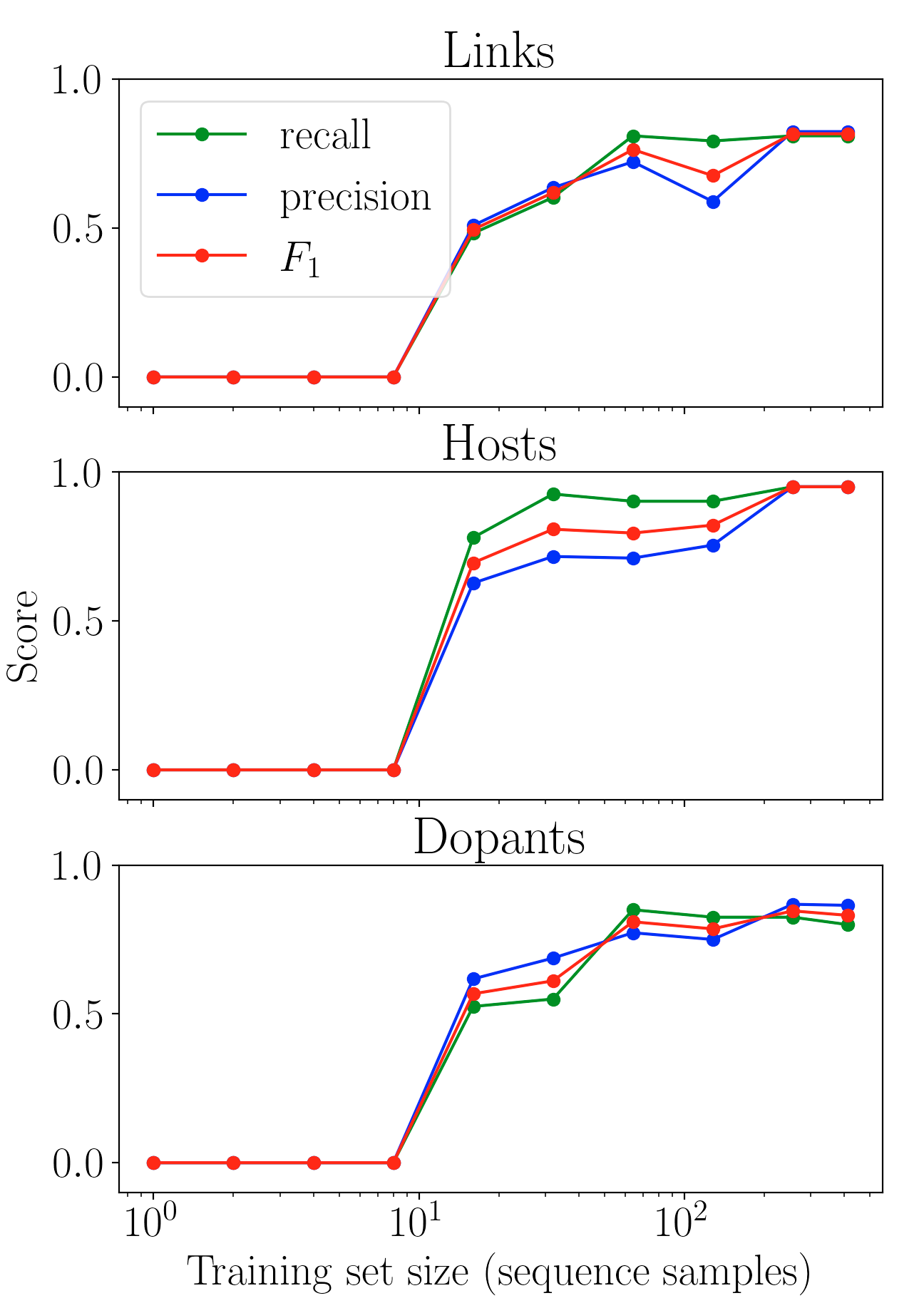}
    \caption{Test set performance by number of training samples for the doping extraction task using the \texttt{Doping-ENG} model.}
    \label{fig:learning-curve}
\end{figure}

The effect of training set size on \texttt{Doping-ENG} test set performance is plotted in Figure \ref{fig:learning-curve} for training set sizes ranging from 1 to 413 text-completion pairs. We observe no parsable output sequences for training set sizes below 8 samples, but there is a sharp increase at 16 samples, potentially due to the inability of the model to produce a parsable output sequence format below a threshold number of samples. Above $64$ samples, we see marginal noisy improvement. Each data point represents a single independent training/prediction experiment, and the internal model variability may explain the score variance at higher levels of training samples.

\section{Discussion}
\label{sec:discussion}

The NERRE scores in Tables \ref{table:results-linking-doping}-\ref{table:manually-computed-f1} provide an overview of raw performance, but the advantages and disadvantages of this method are most easily explained with a discussion on factors not directly shown with $F_1$ scores. The primary advantage of this method is its accessibility and ease-of-use (\hyperref[sec:accessibility]{\textit{Section: Accessibility and Ease-of-use}}), as LLM-NERRE requires only specifying a basic schema, annotating a minimal number of examples, and fine-tuning a model via a publicly available API without extensive NLP knowledge or hyperparameter tuning; the final result is an accurate and extensible model with the ability to extract intricate hierarchies of semantically complex, nuanced, and specialized technical information. Additionally, error correction and normalization may be embedded directly into training examples to reduce the need for postprocessing (\hyperref[sec:implicit]{\textit{Section: Implicit error correction and entity normalization}}). In essence, one can show an LLM-NERRE model both what it should extract \textit{and} how it can be condensed and presented. Finally, we discuss limitations of this approach and directions for future work. We encourage the reader to try the LLM-NERRE method themselves with an interactive demo available at \url{http://www.matscholar.com/info-extraction}.

\subsection*{Accessibility and ease-of-use}\label{sec:accessibility}

In contrast to existing pipeline-based or end-to-end NERRE methods, the LLM-NERRE method may be attractive to domain specialists as it requires little programming experience, NLP expertise, or in-depth postprocessing. Users need only to define a schema for a problem of interest, annotate examples adhering to that schema, and fine-tune a LLM model using publicly-available APIs. The flexibility of the schema also allows for highly intricate and hierarchical knowledge graphs to be extracted directly from scientific text; this is in contrast to methods requiring the enumeration of all possible entity relationships, where success depends more heavily on the tuning of output formats and requires a significant number of training examples for each entity-relation type.

We have found the in-the-loop method shown in Fig. \ref{fig:overview} to greatly decrease annotation time as well. Typically, annotation for scientific information extraction tasks is a tedious and error-prone process. Checking intermediate model outputs for errors is qualitatively easier than creating annotations from scratch. Additionally, using GPT-3 for extraction reduces the number of total training examples needed in comparison with BERT-based models; our NERRE doping models exceed the dedicated MatBERT-NER model in entity recognition precision and $F_1$ despite being trained on far less text, requiring no materials-specific pretraining, and jointly linking entities (vs. MatBERT's NER-only capabilities). As shown in in Figure \ref{fig:learning-curve}, performance of the fine-tuned models generally improves slower as the number of text-completion pairs they have been trained on increases.

\begin{figure}
    \centering
    \includegraphics[width=.8\columnwidth]{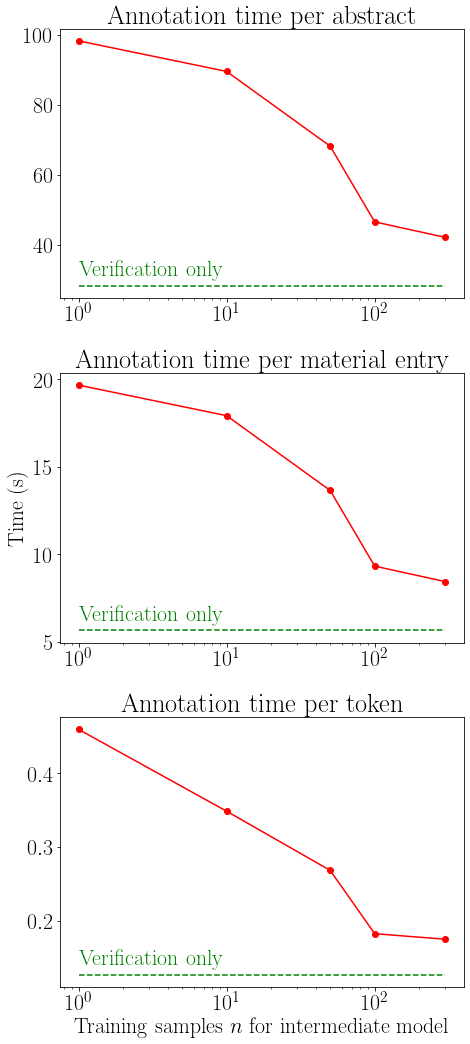}
    \caption{Time taken for a domain expert to annotate new abstracts for the general materials chemistry task with assistance from intermediate (partially-trained) LLM-NERRE models. At each data point, an intermediate model trained on $n$ samples sampled from the original training set (where $n$ is shown on the $x$ axis) infers the completion given the same text prompt presented to the annotator. The pre-populated annotation is then corrected by the annotator. Resulting times for each annotation are shown per abstract, per material mention (JSON document), and per prompt token. The green verification line represents the time taken for the annotator to simply verify whether a given annotation is entirely correct. The verification line represents a lower bound on the annotation time; even with a perfectly-performing model, the annotator must still take time to verify the annotation.}
    \label{fig:annotation-curve}
\end{figure}

As a separate experiment, we compared the usefulness of in-the-loop (error-correction) models for aiding a human domain expert in annotating novel materials-related abstracts in the \texttt{General-JSON} schema. In each trial of the experiment, the human annotator receives 10 novel abstracts and a schema pre-populated with a suggestion from an intermediate version of the \texttt{General-JSON} model trained on $n$ samples of training data ($n=$1, 10, 50, 100, 300). The trained models and the human both receive the same input prompt; the task of the human annotator was to correct the intermediate model's suggestions. The time to complete each annotation was recorded. As shown in Fig. \ref{fig:annotation-curve}, the annotation time sharply decreases as the number of training samples used in the intermediate models increases; the $n=300$ intermediate model was able to reduce the average annotation time per abstract by $57\%$ in comparison with the $n=1$ model. At low numbers of training samples, the models' predictions are un-parsable or essentially useless, and the annotator must create annotations from scratch. At higher numbers of training samples, particularly those above 50, the intermediate model predictions require very little error correction from the annotator. As a lower bound, we also report the time needed by the annotator to simply verify whether an entry was entirely correct (verification time). Hypothetically, a perfect model creating perfect annotations would require the human annotator only to check the outputs rather than correct them. We find that by three metrics (time per abstract, time per material entry, and time per prompt token), the human annotator annotated substantially faster with a well-trained model in the loop ($n$ samples $>50$) than with a poorly trained model ($n$ samples $<50$) or no model. For example, the $n=300$ model reduced the annotation time per token $\sim60\%$ compared to the $n=1$ model and is only 38\% slower than the verification time. Given additional training samples for intermediate models, we expect the annotation to asymptotically approach the verification time. Thus, this method may serve as a useful tool for building even larger benchmark datasets for information extraction tasks.

\subsection*{Implicit error correction and entity normalization}
\label{sec:implicit}

An advantage of the LLM-NERRE method presented here is the ability to automatically correct errors and normalize common entity patterns. While the doping models were trained to extract snippets of text exactly as they appeared in the text prompt, the \texttt{General-JSON} model's training data included simple normalizations and error corrections of entities. For example, the erroneous inclusion of whitespace in chemical formulae is common in the raw abstract text; by including a corrected formula instead of the raw string in the output training sequences, the LLM is capable of resolving the entity to a cleaner form. For example, "Li Co O2" is resolved to "LiCoO2" implicitly without additional post-processing. Similarly, given sufficient training examples, the \texttt{General-JSON} model can resolve text such as "PdO functionalized with platinum" to a normalized form such as \texttt{\{\texttt{formula}: "PdO", \texttt{description}: ["Pt-functionalized"]\}}. These implicit normalization and correction abilities of LLM models may prove useful for domain specialists who desire structured entity formats rather than exact string excerpts directly from the text.

\subsection*{Limitations}

One limitation of our model is that valid output schema formatting is not rigorously enforced in the generation step. The LLM may, for any given sample, output an unparsable sequence. This is particularly apparent when the inference token limit is $<512$ tokens and the schema is JSON, as JSON schema typically requires a larger number of tokens for correct formatting. For example, a nearly-correct output sequence containing 10+ correct entities may be missing a "\}" character and therefore will not be parsable. Outputs are nearly always parsable ($\sim 95\%$ success rate), especially as the number of training examples increases. Failures predominantly occur when the sample exceeds the prompt-completion token limit of GPT-3, which at the time of this work is 2048. Because of this, some abstracts that are too long or too dense with information can not be processed with this method. We observe that this was the case in the vast majority of the unparseable completions where the passage and partial completion exceed the token limit and cut off early before the full completion could be output by the model. 

Another limitation is the tendency of LLMs to generate or invent information that is not actually present in the input text, which is called "hallucination" in LLM literature. However, in some cases a properly conditioned language model may produce useful and correct information that would reasonably be inferred by the reader. This is not possible in strict NERRE models which simply extract entities by labeling words in the original text. However, this means it is also possible for a model to extract the same information from a piece of text in multiple ways. For example, an abstract may mention "p‐ZnSe doped with N" and "nitrogen‐doped ZnSe" in the same passage. Is "doped with N" or "nitrogen-doped" the correct description to extract? Clearly both are correct and either one could be reasonably chosen. Moreover, "N-doped" could also be extracted and would be factually correct even though "N-doped" never occurs in the passage. For this reason, most approaches to information extraction require that the exact phrases used in the original text be extracted, but this can introduce complications in downstream post-processing. One example is the cases where "catalyst", "catalysis", "catalyzed", and "catalyzes" are all extracted as applications for Pt from different source passages, which are more difficult to disambiguate later. For this reason, we decided to prefer normalization of entities during extraction in this work. 

Finally, GPT-3 is currently only accessible through OpenAI's online API. This may present difficulties from several standpoints, as the underlying model is controlled exclusively by a private entity and data must be sent to that entity for processing. Moreover, the cost for inference on large datasets using trained models may be prohibitive. However, given the "black-box" nature of this approach, this prompt/completion engineering method is not exclusive to GPT-3 in principle. We expect open-source $seq2seq$ models of sufficient complexity to be able to reproduce or exceed the results shown here.

\section{Conclusion} \label{sec:conclusion}

We present a facile prompt and completion-engineering method for extracting complex, hierarchical, and highly domain-specific relation information from unstructured scientific text using large $seq2seq$ language models. This method is highly flexible and we expect it can be easily adapted to many problems within scientific domains. In applying this method, we find excellent performance on three diverse tasks for materials engineering: solid-state impurity doping, metal-organic frameworks, and general materials relations. The non-technical nature of this approach implies scientists without NLP training can utilize existing $seq2seq$ models such as GPT-3 to extract large structured relational datasets for highly-specific problems. As the LLM is treated essentially as a black box, we expect this approach may be used LLMs other than GPT-3, including LLMs released in the near future. We hope this approach enables domain specialists to rapidly extract relational datasets for the advancement of scientific knowledge. An online demo is available at \url{http://www.matscholar.com/info-extraction}.

\section{Author Contributions}
J.D., A.D., and N.W. developed the information extraction method presented in this paper and J.D. and A.D. collected the abstract dataset used. J.D originated and performed supporting experiments to justify the sequence-to-sequence approach and use of GPT-3 for document-level information extraction from materials science text. N.W. and J.D. developed the sequence-to-JSON method. A.D. created the doping schemas, developed the sequence-to-sentence method, and annotated the doping dataset. J.D. created the general materials information schema, annotated the general materials dataset, trained the general materials information extraction model, and manually scored the information extraction results for the General-JSON model and the BatteryBERT validation set results. A.D. trained the Doping-JSON, Doping-ENG, models and implemented the MatBERT + Proximity doping model. S.L. trained and collected data for the seq2rel model. A.D. performed the learning curve experiments, and collected data on annotation times. A.S.R. and J.D. co-created the MOF schema and annotated the MOF dataset while J.D. trained the MOF-JSON model. A.J. contributed to task design, task scoring metrics, and overall research directions. G.C., K.P., and A.J. supervised this work. All authors contributed to writing the manuscript.

\section{Acknowledgements}

This work was supported by Toyota Research Institute through the Accelerated Materials Design and Discovery program. A.S.R. acknowledges support via a Miller Research Fellowship from the Miller Institute for Basic Research in Science, University of California, Berkeley. We thank Anna Sackmann (Science Data and Engineering Librarian at UC Berkeley) for helping us to obtain Text and Data Mining agreements with the specified publishers and we also thank J. Montoya and A. Trewartha for helpful discussions. 

\insertbibliography{main}

\clearpage

\section{Supplementary Information}\label{sec:supp}

\renewcommand{\thefigure}{S\arabic{figure}}
\renewcommand{\thetable}{S\arabic{table}}
\setcounter{table}{0}
\setcounter{figure}{0}

\subsection*{Ambiguity of annotations}

In practice, annotation can be a complex task even for trained researchers. We have found qualitatively that the performance of the LLM-NERRE models is limited by how consistently and comprehensively the desired schema and entities can be defined. As opposed to canonical NLP examples, where distinctions between entities such as "person" and "place" are relatively clear, scientific texts often contain entities which could plausibly be considered as one or more kinds of entity depending on definition. Reducing ambiguity in the annotation schema is therefore a source of potential improvement for LLM-NERRE methods. We examine several tasks below in regards to ambiguity.

\textbf{Doping task ambiguity}. Definitional ambiguity is particularly apparent in cases where long-range dopant-host relationships may be referenced with no clear "correct" answer to either the annotator or model. However, most sentence-level dopant-host relationships are very clear to annotators. As mentioned in the main text, this is a primary motivator for using a sentence-level annotation scheme rather than an abstract-level annotation scheme.

\textbf{General task schema ambiguity}. In the \texttt{General-JSON} schema, entity class definitions are not entirely consistent throughout the training set. Class definitions drifted somewhat as annotators found more edge cases during the annotation process, and it was still not possible to completely capture all relevant information contained in a passage with this schema. Keeping these details in mind, annotation was done on a "best effort" basis, which seems to still be sufficient to train models capable of complex information extraction tasks. Additionally, this schema is not currently optimized for composite materials (\textit{e.g.,} "F-TiO2/NO3-layered double hydroxide composite"). We annotated examples such that components of the composite were sometimes listed as separate materials and sometimes listed as a single material. While this partially contributes to lower string-match $F_1$ scores, we find the inconsistencies in the annotations does not significantly impair the model's ability to deal with composites. 

\subsection*{\texttt{Doping-ENG} schema}\label{sec:supp-doping-schema}

The English sequence completion schema for the \texttt{Doping-ENG} and \texttt{DopingExtra-ENG} models are shown below. Each paradigm represents a single line in the output completion; aside from the "No information" paradigm, there may be one or more of each of the paradigms in the output sequence (\textit{e.g.,} there may be multiple separate one-to-many host-dopant relationships, a single dopant with no host, multiple results, and multiple modifiers extracted from the same sentence). The placeholders \texttt{<HOST>}, \texttt{<DOPANT*>}, \texttt{result}, and \texttt{modifier*} are used in place of actual string literal entities. The apostrophe "\texttt{'}" character was used to more easily delimit entity captures. Output sequences not matching one of these patterns were considered not parsable.

\begin{itemize}
    \item \textbf{One host to one-or-more dopants:} \texttt{The host '<HOST>' was doped with '<DOPANT1>', '<DOPANT2>', ... and '<DOPANTN>'.}
    \item \textbf{Single dopant, no host}: \texttt{'<DOPANT>' is a dopant.}
    \item \textbf{Single host, no dopant}: \texttt{The host '<HOST>' was doped.}
    \item \textbf{Single result}: \texttt{'<result>' is a possible doped result formula.}
    \item \textbf{One or more modifiers}: \texttt{Modifiers of the doping are '<modifier1>', '<modifier2>'... '<modifierN>'.} 
    \item \textbf{No doping-related information:} \texttt{There is no doping information}.
\end{itemize}

\subsection*{Sequence reconstruction metrics}

\textbf{Exact Match Accuracy.} The exact sequence match accuracy is defined based on the an exact match between a predicted completion sequence $\hat{c_i}$ and true completion sequence $c_i$, averaged over $n$ all samples to be evaluated:

\begin{equation}
\text{Exact match accuracy} = \frac{\sum_{i}^n \delta_{\hat{c_i}, {c_i}}}{n}
\end{equation} 

\noindent Where $\delta$ is the Kronecker delta:
\begin{equation}
\delta_{\hat{c_i}, {c_i}} = 
\left\{
    \begin{array}{lr}
        0, & \text{if } \hat{c_i} \neq c_i\\
        1, & \text{if } \hat{c_i} = c_i
    \end{array}
\right\}
\end{equation}

\noindent Thus, the exact match accuracy is the most stringent metric, as any character addition (\textit{e.g.,} addition of extra whitespace), missing character, or permutation of entries (\textit{e.g.,} a different ordering of otherwise correct JSON documents within the output) is not an exact match. This is a lower bound on the information capture of the models, as correct information may be equivalently represented multiple ways.

\textbf{Jaro-Winkler Similarity.} For a more granular analysis, we use the Jaro-Winkler similarity, $\Phi$, as a string comparator metric. As in the exact sequence match, we average over all $i=1, 2, 3, ..., n$ evaluation sequences where the predicted string completion is labelled $\hat{c_i}$ and the corresponding true string completion is $c_i$. With weights of the first string, second string, and transposition all set equal, the similarity $\Phi_i$ between a predicted completion string $\hat{c_i}$ and a true completion string $c_i$ is defined as:

\begin{equation}
    \Phi_i = \begin{cases}0, & \text { if } m=0 \\ \frac{1}{3}\left(\frac{m}{\left|\hat{c_i}\right|}+\frac{m}{\left|c_i\right|}+\frac{m-t}{m}\right), & \text { otherwise }\end{cases}
\end{equation}

\noindent Where $m$ is the number of matching characters between $\hat{c_i}$ and $c_i$, $t$ is the number of transpositions, and $|\hat{c_i}|$ and $|c_i|$ are the lengths of the predicted and true completions, respectively. The final average Jaro-Winkler similarity $\Bar{\Phi}$ is the arithmetic mean averaged over $n$ samples, $\Bar{\Phi} = \sum_i^n \Phi_i/n$.

\textbf{Parsability.} As a final sequence reconstruction metric, we report the average percentage of samples which can be parsed from string literal into object form. For the \texttt{*-JSON} models, this indicates that the output sequence is well-formatted JSON; for \texttt{*-ENG} models, this indicates the output sequence adheres to the natural-language like schema on which the model was trained. In either case, the parsability of the output sequences simply indicates whether the sequence can be easily transformed into a relational object. If a sequence can be parsed using the same function to encode training samples, we return a parsability of one; otherwise, parsability is zero. We average the parsability over all $n$ samples of the evaluation dataset to calculate a final average parsability percentage.

\subsection*{Performance on named entity recognition}\label{sec:perform-ner}

\begin{table*}[!htb]
\caption{Exact match (E.M.) named entity recognition scores for the three materials engineering tasks. Highest scores are shown in bold for entities predicted with multiple models (\texttt{host}, \texttt{dopant}).}\label{table:results-recognition}
\centering
    \begin{tabular}{ |c|c|c|c|c|c| } 
    \hline
    Task & Model Name  & Entity & E.M. Recall & E.M. Precision & E.M. $F_1$ \\
    \Xhline{2\arrayrulewidth}
    \multirow{10}{*}{Doping} 
          & \multirow{2}{*}{\texttt{Doping-ENG}} & \texttt{host} & \textbf{0.951} & \textbf{0.928} & \textbf{0.940} \\\cline{3-6}
        & & \texttt{dopant} & 0.775 & 0.837 & 0.805 \\\cline{2-6}
        & \multirow{2}{*}{\texttt{Doping-JSON}} & \texttt{host} & 0.888 & 0.858 & 0.872  \\\cline{3-6}
        & & \texttt{dopant} & 0.714 & 0.750 & 0.732 \\\cline{2-6}
        & \multirow{4}{*}{\texttt{DopingExtra-ENG}} & \texttt{host} & \textbf{0.951} & 0.907 & 0.929\\\cline{3-6}
        & & \texttt{dopant} & \textbf{0.850} & \textbf{0.919} & \textbf{0.883} \\\cline{3-6}
        & & \texttt{results} & 0.736 & 0.933 & 0.824 \\\cline{3-6}
        & & \texttt{modifiers} & 0.500 & 0.333 & 0.400 \\\cline{2-6}
        & \multirow{2}{*}{\texttt{MatBERTDoping} + Proximity} & \texttt{host} & 0.885 & 0.547 & 0.67 \\\cline{3-6}
        & & \texttt{dopant} & 0.727 & 0.600 & 0.658 \\\cline{2-6}
        & \multirow{2}{*}{\texttt{seq2rel}} & \texttt{host} & 0.805 & 0.458 & 0.584 \\\cline{3-6}
        & & \texttt{dopant} & 0.600 & 0.600 & 0.600 \\ 
    \Xhline{2\arrayrulewidth}
    \multirow{6}{*}{General Materials} & \multirow{6}{*}{\texttt{General-JSON}} 
    &   \texttt{acronym} & 0.557 & 0.699 & 0.613 \\\cline{3-6}
    & & \texttt{applications} & 0.700 & 0.732 & 0.715 \\\cline{3-6}
    & & \texttt{name} & 0.588 & 0.778 & 0.668 \\\cline{3-6}
    & & \texttt{formula} & 0.642 & 0.696 & 0.664 \\\cline{3-6}
    & & \texttt{structure\_or\_phase} & 0.581 & 0.639 & 0.608 \\\cline{3-6}
    & & \texttt{description} & 0.484 & 0.507 & 0.494 \\
    \Xhline{2\arrayrulewidth}
    \multirow{5}{*}{MOFs} & \multirow{5}{*}{\texttt{MOF-JSON}} &  \texttt{name} & 0.690 & 0.752 & 0.718\\\cline{3-6}
     & &    \texttt{mof\_formula} & 0.412 & 0.509 & 0.454\\\cline{3-6}
     & &    \texttt{applications} & 0.682 & 0.722 & 0.701\\\cline{3-6}
     & &    \texttt{guest\_species} & 0.614 & 0.754 & 0.666\\\cline{3-6}
     & &    \texttt{description} & 0.386 & 0.544 & 0.446\\\cline{3-6}
    \hline
    \end{tabular}
\end{table*}

We show in Table \ref{table:results-recognition} the named entity recognition scores for recall, precision, and $F_1$ score for each of the models. For the doping task, the highest scores per common entity category (\texttt{host}, \texttt{dopant}) are shown in bold, as these tasks are evaluated with multiple models (\texttt{Doping-ENG}, \texttt{Doping-JSON}, and \texttt{DopingExtra-ENG}). Entities are evaluated on an exact per-word match basis rather than the basis of exact matches between entire multi-word entities, as it is unclear even to annotators exactly where to denote the end of some complex multi-word entities (e.g., "ZnO nanoparticle crystals" vs. "ZnO nanoparticle" vs. "ZnO"). However, since compositions are the core part of the desired data for all tasks, all words of any formula entity (and hence all its links) are marked incorrect if the entire composition is not captured exactly. Support for each class may be found in Supplementary Section \ref{supp-support}.

In the doping task, we generally observe the highest scores for host recognition from the \texttt{Doping-ENG} model and the highest scores for dopant recognition from the \texttt{DopingExtra-ENG} model. $F_1$ scores for the \texttt{Doping-JSON} model are 7-9\% lower than those of the \texttt{Doping-ENG} model. Between the \texttt{DopingExtra-ENG} and \texttt{Doping-ENG} models, we find a 1\% difference in \texttt{host} $F_1$ in favor of the simpler \texttt{Doping-ENG} model but 8\% higher \texttt{dopant} $F_1$ in favor of the \texttt{DopingExtra-ENG} model. This is a notable result as the \texttt{DopingExtra-ENG} model must additionally extract \texttt{results} and \texttt{modifiers} entities in the output sequence. We speculate that this discrepancy is due to a combination of model variability and extra specification of non-dopant chemical species within the \texttt{DopingExtra-ENG} model's training set. For example, when the \texttt{DopingExtra-ENG} model is trained with "high-doping" as a labelled \texttt{modifier}, it may learn "high" is not a valid dopant entry. This may be useful in the design of more complex completion schema, as \texttt{DopingExtra-ENG} is an example of a \textit{more} complex completion schema extracting all of its entities more accurately.

\subsection*{GPT-3 inference parameters}

All doping models were trained with 7 epochs. Intermediate models shown in Figure \ref{fig:learning-curve} were trained with a number of epochs depending on the number of training samples $t$: 2 epochs for $2^0 \leq t < 2^6$, 4 epochs for $2^6 \leq t \leq 2^7$, and 7 epochs for $t = 2^8$. The remaining models were trained with a batch size of 1 for 4 epochs with start sequence \texttt{"\textbackslash n\textbackslash n\textbackslash n\#\#\#\textbackslash n\textbackslash n\textbackslash n"} and stop sequence \texttt{"\textbackslash n\textbackslash n\textbackslash nEND\textbackslash n\textbackslash n\textbackslash n"}. All doping models used GPT-3 inference parameters of 256 max tokens, 0 temperature, \texttt{"\textbackslash n\textbackslash n\textbackslash n\#\#\#\textbackslash n\textbackslash n\textbackslash n"} start token, and \texttt{"\textbackslash n\textbackslash n\textbackslash nEND\textbackslash n\textbackslash n\textbackslash n} end token. All other models used 512 max tokens with remaining parameters identical to the doping models.

\subsection*{seq2rel parameters}
\label{supp:seq2rel-params}

 Seq2rel models were trained using 267 doping sentences that had dopant-basemat links, with 4 different training:validation splits (90:10, 80:20, 70:30, 95:5) and the model with the highest validation micro-F1 score was selected. Training configurations for seq2rel used parameters for training gene-disease association (GDA) used in Giorgi et al.\cite{seq2rel}, while modifying the entity tokens to \texttt{"@DOPANT@","@BASEMAT@"} and relation token to \texttt{"@DBR@"}. To be specific:

\begin{verbatim}
model_name = microsoft/BiomedNLP-PubMedBERT-
    base-uncased-abstract-fulltext
max_length=512
max_steps=96
num_epochs=30
batch_size=1
grac_acc_steps=1
decoder_lr=5e-4
encoder_lr=2e-5
encoder_wd=0.01
reinit_layers=1
weight_dropout=0.5
beam_size=4
length_penalty=0.8
\end{verbatim}

\subsection*{Doping task graph example}

\begin{figure}[h!]
    \centering
    \includegraphics[width=\columnwidth]{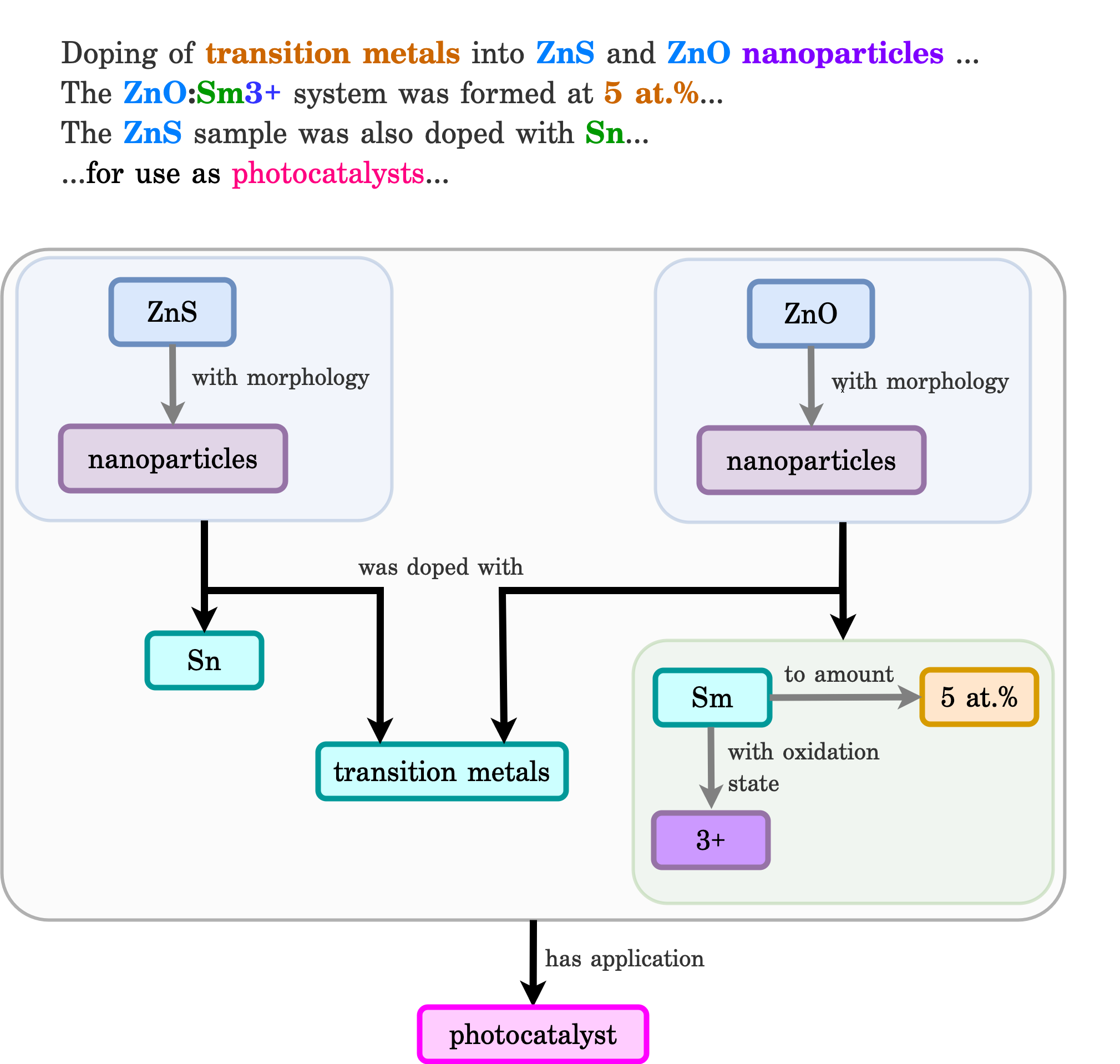}
    \caption{An example complex graph resolved from the outputs of a LLM-NERRE model. The distinction of this graph from typical entity relationship graphs is the hierarchical format. This hierarchical graph, in contrast to a flat graph, may denote that \ch{ZnO} \textit{nanoparticles} - as opposed to bulk wurtzite \ch{ZnO} - were doped by representing an entity with its own subgraph. Similarly, the samarium dopant is represented as a subgraph specifying its oxidation state and amount. Finally, all dopant relationships for \ch{ZnO} and \ch{ZnS} are linked to an application "photocatalyst". This hierarchical graph specifies a much more precise series of relationships extracted from text; for example, \textit{"ZnO (as nanoparticles) was doped with Sm (having an oxidation state of +3, and to an amount of 5 atomic percent) resulting in a photocatalyst."} is more precise than a flat-graph relationship such as \textit{"ZnO is a photocatalyst"}.}
    \label{fig:supp-doping-graph}
\end{figure}

In Figure \ref{fig:supp-doping-graph} an example end result from a LLM-NERRE model is shown in graph format. We aim here to recognize not just relationships between individual entities, but hierarchical relationships with relationship types which need not be explicitly and comprehensively enumerated beforehand. The LLM-NERRE models presented in the main text provide a step towards this kind of final product.

\subsection*{Class support for materials extraction tasks}
\label{supp-support}

    \begin{table}[H]
    \caption{Class support for doping tasks among 77 test (gold) set sentences. \label{table:supp-support-doping}}
    
    \centering
  \begin{tabular}{ |l|c| } 
         \hline
         Class/Relation & Support \\
        \hline
        \texttt{host} & 76 \\
        \hline
        \texttt{dopant} & 60 \\
        \hline
        \texttt{host} $\rightarrow$ \texttt{dopant} (link) & 72 \\
        \hline
        \texttt{results} & 12 \\
        \hline
        \texttt{modifiers} & 7 \\
        \hline
  \end{tabular}
\end{table}

    \begin{table}[H]
    \caption{Class support for the general materials NERRE task from 320 total test set abstracts, including the core entities and the subset of potential links shown in results. \label{table:supp-support-general}}
    \centering
  \begin{tabular}{ |l|c| } 
         \hline
         Class/Relation & Support \\
        \hline
\texttt{acronym} & 61 \\
\hline
\texttt{applications} & 527 \\
\hline
\texttt{name} & 192 \\
\hline
\texttt{formula} & 386 \\
\hline
\texttt{structure\_or\_phase} & 402 \\
\hline
\texttt{description} & 288 \\
\hline
\texttt{formula} $\rightarrow$ \texttt{acronym} (link) & 14 \\
\hline
\texttt{formula} $\rightarrow$ \texttt{application} (link) & 49 \\
\hline
\texttt{formula} $\rightarrow$ \texttt{name} (link) & 49 \\
\hline
\texttt{formula} $\rightarrow$ \texttt{structure\_or\_phase} (link) & 368 \\
\hline
\texttt{formula} $\rightarrow$ \texttt{description} (link) & 199 \\
\hline
  \end{tabular}
\end{table}

\begin{table}[H]
    \caption{Class support for the MOF task from 255 total test set abstracts including the core entities and the subset of potential links shown in results. \label{table:supp-support-mofs}}
    
\centering
  \begin{tabular}{ |l|c| } 
           \hline
         Class/Relation & Support \\
        \hline
\texttt{application} & 953 \\
           \hline
\texttt{guest\_species} & 205 \\
           \hline
\texttt{description} & 178 \\
           \hline
\texttt{name} & 434 \\
           \hline
\texttt{mof\_formula} & 91 \\
           \hline
\texttt{mof\_formula} $\rightarrow$ \texttt{application} (link) & 171 \\
           \hline
\texttt{mof\_formula} $\rightarrow$ \texttt{guest\_species} (link) & 21 \\
           \hline
\texttt{mof\_formula} $\rightarrow$ \texttt{description} (link) & 44 \\
           \hline
\texttt{mof\_formula} $\rightarrow$ \texttt{name} (link) & 29 \\
        \hline
  \end{tabular}
\end{table}

\end{document}